\documentclass[preprint,12pt]{elsarticle}

\usepackage{amssymb}
\usepackage{booktabs}
\usepackage{multirow}
\usepackage{amsmath}
\usepackage{subfigure}
\usepackage{algorithmic}
\usepackage{algorithm}
\usepackage{url}
\usepackage{array}
\usepackage{threeparttable}
\def\etal{\emph{et al.}}

\journal{elsarticle}

\begin{document}

\begin{frontmatter}


\title{RADAP: A Robust and Adaptive Defense Against Diverse Adversarial Patches on Face Recognition}




\author[1,2]{Xiaoliang Liu}
\ead{xiaoliang_liu@smail.nju.edu.cn}
\author[1,3]{Furao Shen}
\ead{frshen@nju.edu.cn}
\author[4]{Jian Zhao}
\ead{jianzhao@nju.edu.cn}
\author[1,2]{Changhai Nie}
\ead{changhainie@nju.edu.cn}
\address[1]{State Key Laboratory for Novel Software Technology, Nanjing University, China}
\address[2]{Department of Computer Science and Technology, Nanjing University, China}
\address[3]{School of Artificial Intelligence, Nanjing University, China}
\address[4]{School of Electronic Science and Engineering, Nanjing University, China}

\begin{abstract}
Face recognition (FR) systems powered by deep learning have become widely used in various applications. However, they are vulnerable to adversarial attacks, especially those based on local adversarial patches that can be physically applied to real-world objects. In this paper, we propose RADAP, a robust and adaptive defense mechanism against diverse adversarial patches in both closed-set and open-set FR systems. RADAP employs innovative techniques, such as FCutout and F-patch, which use Fourier space sampling masks to improve the occlusion robustness of the FR model and the performance of the patch segmenter. Moreover, we introduce an edge-aware binary cross-entropy (EBCE) loss function to enhance the accuracy of patch detection. We also present the split and fill (SAF) strategy, which is designed to counter the vulnerability of the patch segmenter to complete white-box adaptive attacks. We conduct comprehensive experiments to validate the effectiveness of RADAP, which shows significant improvements in defense performance against various adversarial patches, while maintaining clean accuracy higher than that of the undefended Vanilla model.
\end{abstract}






\end{frontmatter}


\section{Introduction}
\label{sec:introduction}
Face recognition (FR) technology is used extensively in our everyday lives, from unlocking our phones to verifying payments and identifying individuals. With the advancement of deep learning, FR has achieved remarkable progress in recent years~\cite{deng2019arcface,chen2018mobilefacenets,meng2021magface}. However, deep neural networks are also vulnerable to adversarial attacks, which can fool FR systems by adding imperceptible perturbations to the input images~\cite{goodfellow2014explaining,madry2017towards}. Among various types of adversarial attacks, local-based adversarial patches are especially threatening, as they can be printed and attached to physical objects in the real world, such as glasses~\cite{sharif2016accessorize}, hats~\cite{komkov2021advhat}, or masks~\cite{liu2022rstam,zolfi2022adversarial}. These patches may lead to misclassification of the individual who wears them or exclusion of other people in the scene by the FR system. The physical realizability of adversarial patches poses a greater threat to real-world FR systems than the global perturbation-based adversarial examples. Therefore, in this paper, we focus on how to defend against adversarial patches on FR.

FR systems can be categorized as either closed-set or open-set systems. Closed-set systems aim to identify individuals from a predefined set of known subjects, while open-set systems are designed to accommodate recognition of unknown individuals not present in the predefined set. Closed-set FR can be compared to a problem of image classification in which the goal is to categorize a face within specific and predefined categories. In contrast, open-set FR is akin to a distance metric problem, which involves the measurement of similarity between the query face and the known subjects.

While many adversarial patch defense methods primarily focus on image classification tasks, such as Interval Bound Propagation (IBP)~\cite{chiang2020certified}, Clipped BagNet (CBN)~\cite{zhang2020clipped}, Defense against Occlusion Attacks (DOA)~\cite{wu2020defending}, Local Gradients Smoothing (LGS)~\cite{naseer2019local}, PatchGuard (PG)~\cite{xiang2021patchguard}, PatchCleanser (PC)~\cite{xiang2022patchcleanser}, it is crucial to recognize that in the real world, FR systems often operate as open-set systems. This is because they must contend with the possibility of encountering unknown individuals beyond their training data when deployed in uncontrolled environments. 

\textbf{As a result, the first challenge in developing effective defense strategies is addressing both patch attacks in closed-set FR systems and open-set FR systems.} To address this challenge, we utilize a patch segmenter capable of detecting adversarial patches and subsequently concealing them using masking techniques. This process introduces four main challenges.

\textbf{Firstly, there is a need to acquire suitable training data for the patch segmenter and ensure its capability to detect adversarial patches of various shapes.} To tackle this issue, we present F-patch, a method that generates patches with diverse shapes through random sampling in Fourier space. This approach equips our trained patch segmenter to recognize a broad spectrum of adversarial patches.

\textbf{Secondly, we must enhance the patch segmenter's ability to detect patch boundaries.} To further refine the accuracy of patch detection, we propose an edge-aware binary cross-entropy (EBCE) loss function aimed at improving edge perception.

\textbf{Thirdly, as we remove adversarial patches by applying masks, it becomes imperative to fortify the occlusion robustness of the FR model.} Leveraging the well-established Cutout data augmentation technique, we introduce a mask created through random Fourier space sampling, which we term FCutout. This augmentation method significantly strengthens the model's capacity to handle occlusions.

\textbf{Lastly, given that our patch segmenter employs deep neural networks, we must address the vulnerability to complete white-box adaptive attacks after the patch segmenter model is leaked.} To mitigate this issue, we propose the split and fill (SAF) strategy, which substantially enhances the robustness of our defense method. 

Our contributions can be summarized as follows:
\begin{itemize}
\item We introduce RADAP, a robust and adaptive defense approach designed to counter diverse adversarial patches in both closed-set and open-set FR systems.

\item To enhance the occlusion robustness of the FR model and optimize the performance of the patch segmenter, we propose two novel techniques: FCutout and F-patch. These methods employ randomly sampled masks in Fourier space. Additionally, we introduce the EBCE loss function to further enhance the patch segmenter's performance.

\item We present the SAF strategy, specifically developed to mitigate the risk of complete white-box adaptive attacks following the leakage of the segmenter model.

\item Ultimately, we validate the efficacy of RADAP through a series of extensive experiments. Comparative analysis with other state-of-the-art methods demonstrates RADAP's significant improvements in defense performance against a wide range of diverse adversarial patches. Notably, the clean accuracy of RADAP surpasses that of the undefended Vanilla model.
\end{itemize}

\section{Related Works}
\label{sec:related_works}
This section presents a review of pertinent literature concerning Adversarial Patches in the context of FR and the strategies employed for Defending Against Adversarial Patches.
\subsection{Adversarial Patches on Face Recognition}
The primary objective of adversarial patch attacks on FR systems is to manipulate the decision-making process, potentially leading to misclassifications or unauthorized access. These attacks can be broadly categorized into two main types: evasion attacks and impersonation attacks. Evasion attacks focus on reducing the confidence in similarity measurements between identical identity pairs, resulting in misclassifications and a degradation of system accuracy. In contrast, impersonation attacks aim to enhance the feature similarity between the source images of the target individual and the various identities used by the attacker, thereby enabling unauthorized access to the FR system.

A substantial number of methods for adversarial patch attacks have been proposed, including Adv-Glasses~\cite{sharif2016accessorize}, Adv-Hat~\cite{komkov2021advhat}, and Adv-Masks~\cite{zolfi2022adversarial}. These methods ingeniously employ adversarial patches such as eyeglass frames, hats, and respirators, which have proven to be remarkably effective in real-world scenarios. Moreover, the transferability of adversarial patches has been significantly improved through methods like TAP~\cite{xiao2021improving} and RSTAM~\cite{liu2022rstam}, which diversify the inputs. TAP achieves this by generating patches with a variety of attributes and styles, allowing them to adapt to different facial features and environmental scenarios. In contrast, RSTAM leverages random similarity transformations to accommodate variations in facial poses and lighting conditions.

Collectively, these studies underscore the vulnerabilities inherent in facial recognition systems and emphasize the critical importance of our research on developing defenses against adversarial patches.

\subsection{Defense Against Adversarial Patches}
Numerous researchers have dedicated their efforts to defending against adversarial patches. Among these, JPEG Compression (JPEG)~\cite{dziugaite2016study}, Digital Watermark (DW)~\cite{hayes2018visible}, and LGS~\cite{naseer2019local} emerged as early solutions. JPEG compression alleviates the impact of adversarial patches through image compression. DW identifies unnaturally dense regions in the classifier's significance map and conceals them to prevent their influence on classification. LGS is based on the observation that patch attacks introduce concentrated high-frequency noise and proposes gradient smoothing for areas with high gradient magnitude.

However, these methods are susceptible to complete white-box adaptive attacks in the event of defense method compromise. In response to this challenge, Jiang \etal introduced IBP~\cite{chiang2020certified}. Similarly, the adversarial training-based method DOA~\cite{wu2020defending} does not suffer from this vulnerability. More recently, Xiang \etal put forth PG~\cite{xiang2021patchguard}, a network characterized by small receptive fields and outlier masking. However, it requires substantial modifications to the backbone classifier. In recognition of this, Xiang \etal also introduced PC~\cite{xiang2022patchcleanser}, a two-stage judgment-based approach applicable to any classifier. In contrast to the judgment-based PC, Segment and Complete (SAC)~\cite{liu2022segment} deploys a deep neural network for adversarial patch detection and removal.

While PG and PC excel at defending against small patches, they face challenges when confronted with larger patches. To enhance defenses against adaptive attacks, SAC employs completion techniques, albeit rendering it less effective against random multiple patches. In contrast, our approach, RADAP, overcomes these limitations, providing adaptive defense against a wide array of adversarial patches.

\begin{figure*}[]
    \centering
    \includegraphics[width=0.8\linewidth]{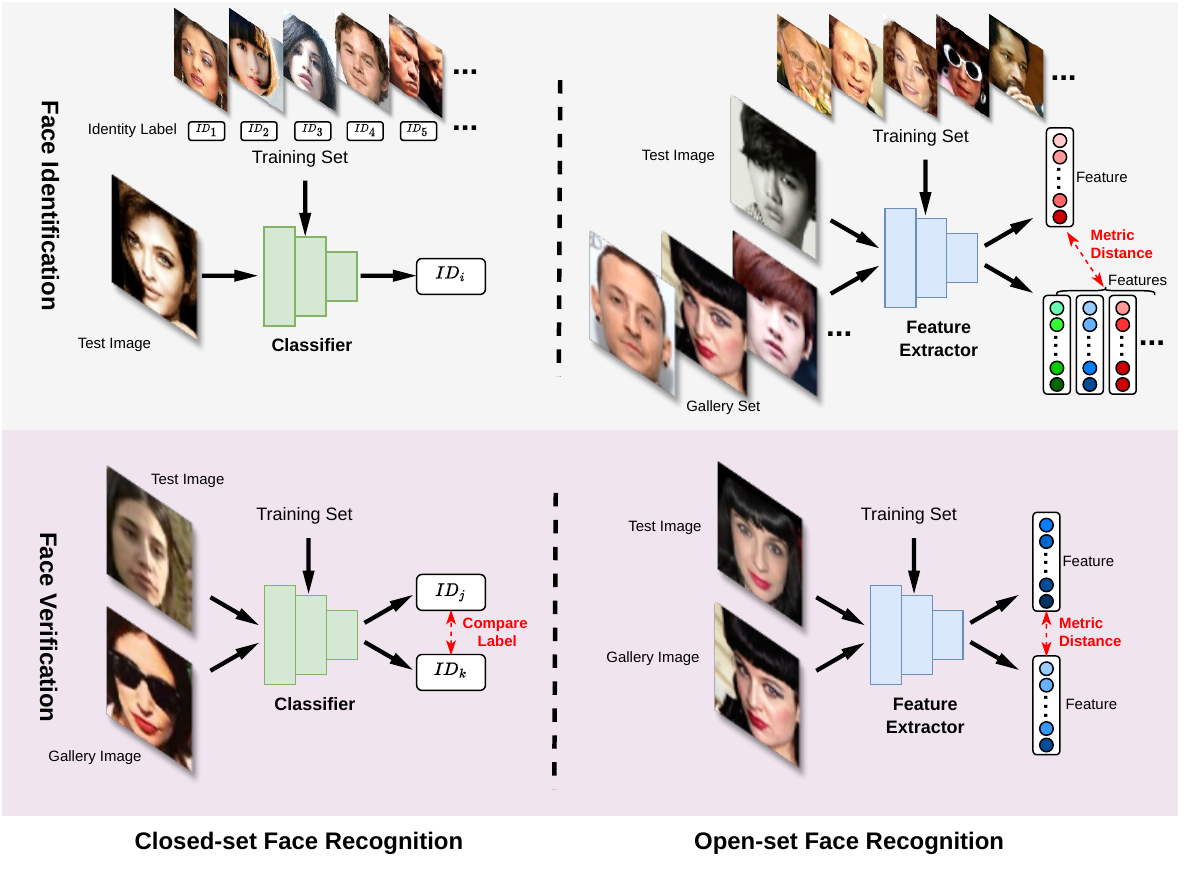}
    \caption{Comparison of closed-set and open-set face recognition.}
    \label{fig:fr}
\end{figure*}
\section{Preliminary}
\label{sec:preliminary}
\subsection{Face Recognition}
In this paper, we focus on defense against adversarial patches on FR systems. We begin with a brief overview of FR systems and their tasks. As depicted in Figure~\ref{fig:fr}, FR encompasses two primary tasks: face identification and face verification. Face identification is the task of identifying a specific individual in an image by comparing it to known faces. On the other hand, face verification determines whether two given images belong to the same person. FR can be further divided into two categories: closed-set and open-set FR. In closed-set recognition, the system works with predefined identities, treating face identification as a multi-class classification problem. Neural networks used for closed-set FR are generally considered as classifiers in this context. In contrast, open-set recognition deals with unknown identities and employs metric learning techniques to measure the similarity between images in a selected feature space during face verification. Open-set face identification takes this a step further by comparing distances between the test image and each gallery image to identify the closest match. In open-set recognition, neural networks are often viewed as feature extractors. 

\begin{table}[]
\centering
\caption{Summary of important notation}
\scalebox{1.0}{
\begin{tabular}{@{}ll@{}}
\toprule
Notation & Description  \\ \midrule
$\boldsymbol{x}_{a}$    &   A attack image. \\
$\boldsymbol{x}_{s}$    &   A attacker’s source image.\\
$\boldsymbol{x}_{t}$    &   A target image.  \\
${y}_{s}$    &   A identity label of $\boldsymbol{x}_{s}$.\\
${y}_{t}$    &   A identity label of $\boldsymbol{x}_{t}$. \\
$\boldsymbol{M}$ & A binary mask. \\
$\odot$ & The element-wise product operator. \\ 
$\mathcal{F}_{FR}$ & A face recognition model. \\
$\mathcal{L}_{CE}$ &  The cross-entropy loss function. \\
$\mathcal{D}_{CS}$ &  The cosine similarity distance measure function. \\
$\boldsymbol{X}_{ij}$  & The $i,j$th element of the two dimensional matrix $\boldsymbol{X}$. \\
$\otimes$ & The convolutional product operator.\\
\bottomrule
\end{tabular}
}
\label{tab:notation}
\end{table}

\begin{table}[]
\centering
\caption{Attack formulation}
\begin{threeparttable}
\scalebox{1.0}{
\begin{tabular}{@{}l|l|l@{}}
\toprule
Target System               & Attacker’s Goal & Formulation \\ \midrule
\multirow{2}{*}{Closed-set} & Evasion         &  $\mathop{\operatorname{argmax}} \limits_{\boldsymbol{x}_a} \mathcal{L}_{CE}(\mathcal{F}_{FR}(\boldsymbol{x}_a),y_s)$           \\ \cmidrule(l){2-3}
                            & Impersonation   &  $\mathop{\operatorname{argmin}} \limits_{\boldsymbol{x}_a} \mathcal{L}_{CE}(\mathcal{F}_{FR}(\boldsymbol{x}_a),y_t)$           \\ \midrule
\multirow{2}{*}{Open-set}   & Evasion         & $\mathop{\operatorname{argmax}} \limits_{\boldsymbol{x}_a} \mathcal{D}_{CS}(\mathcal{F}_{FR}(\boldsymbol{x}_a),\mathcal{F}_{FR}(\boldsymbol{x}_s))$            \\ \cmidrule(l){2-3}
                            & Impersonation   &    $\mathop{\operatorname{argmin}} \limits_{\boldsymbol{x}_a} \mathcal{D}_{CS}(\mathcal{F}_{FR}(\boldsymbol{x}_a),\mathcal{F}_{FR}(\boldsymbol{x}_t))$         \\ \bottomrule
\end{tabular}
}
\begin{tablenotes}
\item[*] $\text{s.t.}\ \boldsymbol{x}_a \odot (1-\boldsymbol{M}) = \boldsymbol{x}_s \odot (1-\boldsymbol{M})$
\end{tablenotes}
\end{threeparttable}
\label{tab:af}
\end{table}

\subsection{Attack Formulation}
This paper aims to address a range of challenges in the field of FR, such as adversarial patch attacks in both closed and open set scenarios, as well as evasion and impersonation attacks. To delineate the adversarial patch attacks, we introduce essential concepts and variables. Specifically, we denote the attacker's source image as $\boldsymbol{x}_s$, the target image as $\boldsymbol{x}_t$, and the attack image as $\boldsymbol{x}_a$. Additionally, we utilize the pre-trained FR model, represented as $\mathcal{F}_{FR}$, and a binary mask, denoted as $\boldsymbol{M}$. Furthermore, we define the cross-entropy loss function, denoted as $\mathcal{L}_{CE}$, as follows:
\begin{equation}
    \mathcal{L}_{CE}(p,y)=-\sum_{i=1}^{K} y_i\log(p_i),
\end{equation}
where $K$ represents the number of classes, $y_i$ denotes the true label for class $i$, and $p_i$ indicates the predicted probability for class $i$. Additionally, we introduce the cosine similarity distance measure function, denoted as $\mathcal{D}_{CS}$:
\begin{equation}
    \mathcal{D}_{CS}(\boldsymbol{v}_1,\boldsymbol{v}_2)= \frac{<\boldsymbol{v}_1,\boldsymbol{v}_2>}{\|\boldsymbol{v}_1\|\|\boldsymbol{v}_2\|},
\end{equation}
where $\boldsymbol{v}_1$ and $\boldsymbol{v}_2$ denote two vectors, and $<\cdot,\cdot>$ signifies the inner product of the vectors.

For clarity, we provide a summary of essential notations in Table~\ref{tab:notation}. Additionally, we outline the process of generating adversarial patches as optimization problems in Table~\ref{tab:af}.

\section{Method}
\label{sec:method}
In this section, we introduce our defense method for FR, which is named "Robust and Adaptive Defense Against Diverse Adversarial Patches" (RADAP). The core concept behind our approach revolves around identifying the locations of adversarial patches through patch segmentation and subsequently eliminating (occluding) these patches. This process necessitates solutions to four key challenges:
\begin{itemize}
\item Enhancing the occlusion robustness of the FR model.
\item Devising a methodology for generating the adversarial patches utilized in training the patch segmenter.
\item Elevating the segmentation accuracy of the patch segmenter.
\item Addressing security concerns associated with the patch segmenter, especially in scenarios where it might leak information.
\end{itemize}

We outline the four fundamental steps of our method, each of which corresponds to one of the aforementioned challenges:
\begin{itemize}
\item \textbf{Step 1:} We train the FR model using the Fourier space sampling-based Cutout (FCutout) data augmentation method to bolster occlusion robustness (see Section~\ref{fcutout}).
\item \textbf{Step 2:} Leveraging the FR model trained in Step 1, we generate Fourier space sampling-based adversarial patch (F-patch) images, which will be used to train the patch segmenter (see Section~\ref{fpatch}).
\item \textbf{Step 3:} We train a patch segmenter model with the F-patch images obtained in Step 2, employing the edge-aware binary cross-entropy (EBCE) loss function, which can improve the segmentation accuracy. Using the trained patch segmenter, we can generate an initial patch mask for adversarial patch segmentation (see Section~\ref{ebce}). 
\item \textbf{Step 4:} To further fortify the security of our approach and mitigate the risk of information leakage from the patch segmenter model, we propose the split and fill (SAF) strategy to manipulate the initial patch masks (see Section~\ref{saf}).
\end{itemize}

FCutout, F-patch, EBCE, and SAF are proposed to address the four key challenges mentioned above. To ensure a comprehensive understanding of our methodology, we introduce the Fourier space sampling-based mask (F-mask) before delving into the four key steps (see Section~\ref{fmask}).
\subsection{Fourier Space Sampling-based Mask}
\label{fmask}
We initiate by introducing the F-mask generation process, a technique derived from FMix~\cite{harris2020fmix}. This process commences with the sampling of a low-frequency grayscale mask from the Fourier space.  Subsequently, the gray-scale mask undergoes a transformation into a binary form through a thresholding mechanism. To elaborate further, let us consider $\boldsymbol{Z}$ as a complex random variable defined within the domain $\mathbb{C}^{H \times W}$. Moreover, both $\mathcal{R}(\boldsymbol{Z})$ and $\mathcal{J}(\boldsymbol{Z})$  follow a normal distribution $\mathcal{N}(\boldsymbol{0},\boldsymbol{I}_{H \times W})$, where $\mathcal{R}$ and $\mathcal{J}$ denote the extraction of real and imaginary parts, respectively.

Following this, we apply a low-pass filter, denoted as $\mathcal{F}_{LPF}$, to $\boldsymbol{Z}$. This filter serves to attenuate the high-frequency components of $\boldsymbol{Z}$ using a decay power parameter $\delta$. Mathematically, this attenuation can be expressed as:

\begin{equation}
\mathcal{F}_{LPF}(\boldsymbol{Z},\delta)_{ij} = \frac{\boldsymbol{Z}_{ij}}{\mathcal{F}_{Freq}(H, W)_{ij}^{\delta}},
\end{equation}
where $\mathcal{F}_{Freq}$ returns the discrete Fourier transform sample frequencies. By applying the inverse discrete Fourier transform $\mathcal{F}_{IDFT}$ to $\mathcal{F}_{LPF}(\boldsymbol{Z},\delta)$, we obtain a gray-scale image represented by:

\begin{equation}
\boldsymbol{G} = \mathcal{F}_{Norm}(\mathcal{R}(\mathcal{F}_{IDFT}(\mathcal{F}_{LPF}(\boldsymbol{Z},\delta)))),
\end{equation}
where $\mathcal{F}_{Norm}$ is a normalisation function, and $\mathcal{R}$ returns the real part of the input. 

The gray-scale mask, $\boldsymbol{G}$, can then be transformed into a binary mask, $\boldsymbol{M}$, using a thresholding function $ \mathcal{F}_{T}$. Let $\mathcal{F}_{T}([\lambda HW], \boldsymbol{G})$ denotes taking the $[\lambda HW]$-th largest value in $\boldsymbol{G}$, where $[\cdot]$ denotes the rounding operation. We can obtain a binary mask represented by
\begin{equation}
\begin{aligned}
&\boldsymbol{M}_{ij}= \begin{cases}1, & \text { if } \boldsymbol{G}_{ij} > \mathcal{F}_{T}([\lambda HW], \boldsymbol{G}), \\ 0, & \text { otherwise,}\end{cases}\\
&\lambda \sim \mathcal{U}(a,b), 0\leq a<b \leq 1,
\end{aligned}
\end{equation}
where $\mathcal{U}(a,b)$ denotes the uniform distribution of $a$ to $b$.

\begin{figure*}[]
    \centering
    \includegraphics[width=0.9\linewidth]{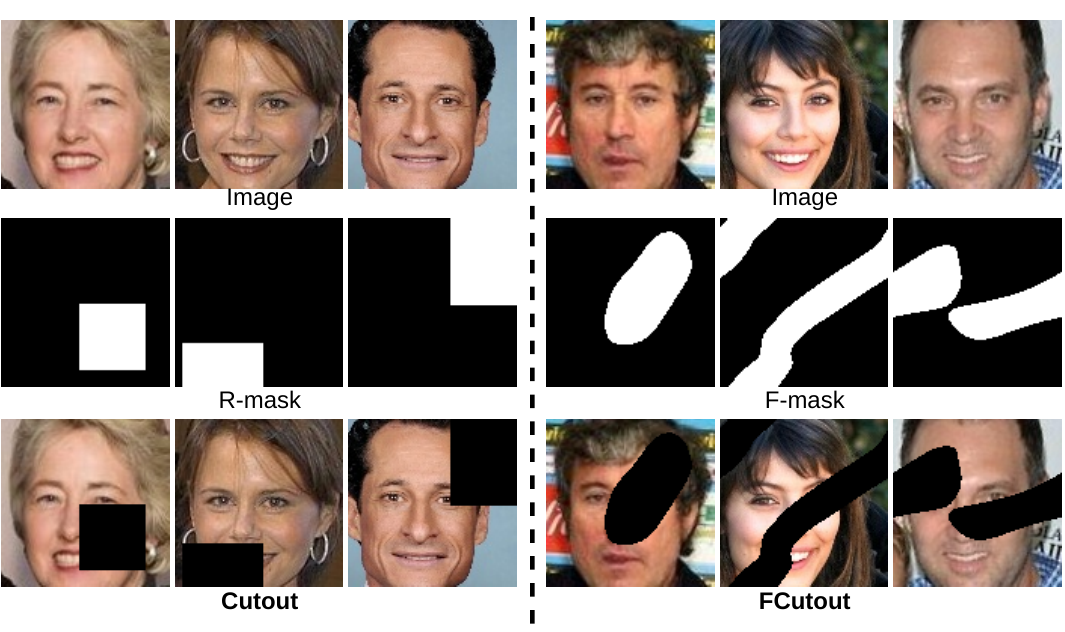}
    \caption{Comparison of Cutout and FCutout.}
    \label{fig:fcutout}
\end{figure*}
\subsection{Step 1: Train the FR Model}
\label{fcutout}
In this step, we train the FR model using FCutout. Our objective is to defend against adversarial patch attacks by removing (occluding) these patches, necessitating the use of occlusion-robust FR models. The Cutout data augmentation method~\cite{devries2017improved} is known to enhance the occlusion robustness of deep neural network models. However, the random rectangular mask (R-mask) employed by Cutout tends to be too uniform, whereas adversarial patch shapes exhibit diversity, resulting in various occlusion patch shapes. To address this limitation, we introduce the FCutout data augmentation method, which leverages the F-mask. Figure~\ref{fig:fcutout} illustrates a comparison between FCutout and Cutout. Notably, the F-mask, unlike the R-mask, offers a wider variety of masks. This diversity results in improved occlusion robustness in FR models trained using FCutout.

\begin{figure*}[]
    \centering
    \includegraphics[width=0.9\linewidth]{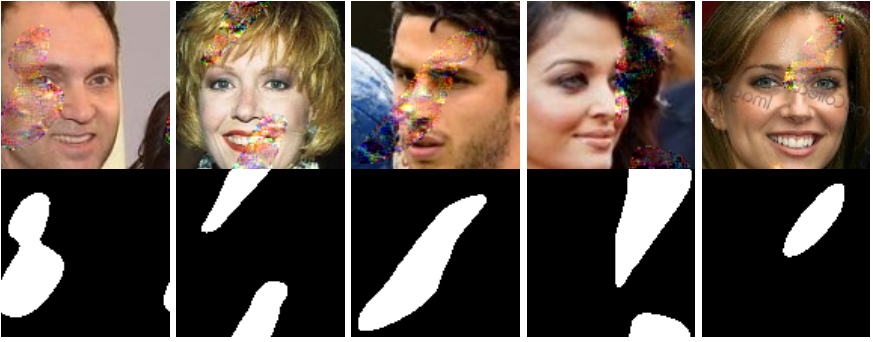}
    \caption{Examples of F-patch.}
    \label{fig:fpatch}
\end{figure*}
\subsection{Step 2: Generate F-patch}
\label{fpatch}
In this step, we generate the adversarial patches required for training the patch segmenter. To ensure diversity in these adversarial patches, we also incorporate the use of the F-mask. Let $\boldsymbol{M}$ denote the binary mask generated using the F-mask, $\boldsymbol{x}_s$ represent the attacker's source image, where $\boldsymbol{x}_s \in [0,1]^{H \times W \times 3}$. Furthermore, $\mathcal{F}_{FR}$ denotes the FR model trained in Step 1. We employ the projected gradient descent (PGD)~\cite{madry2017towards} to generate the attack image $\boldsymbol{x}_a$ with F-patch, as described below:

\begin{equation}
    \begin{aligned}
          \boldsymbol{x}_{a}^{(0)} & = \boldsymbol{x}_s,\\
          \boldsymbol{x}_{a}^{(t+1)} & = \prod_{\mathbb{X}_a}(\boldsymbol{x}_a^{(t)}+\boldsymbol{M}\odot(\alpha\mathop{\operatorname{Sign}}(\nabla_{\boldsymbol{x}_{a}^{(t)}}\mathcal{L}_{CE}(\mathcal{F}_{FR}(\boldsymbol{x}_a^{(t)}),y_s)))),\\
         \boldsymbol{x}_a & = \boldsymbol{x}_a^{(T)}, T \sim \mathcal{U}(1,1000),
    \end{aligned}
\end{equation}
where $\alpha$ represents the perturbation step size, $T$ is the iteration number. To further generating diversity $\boldsymbol{x}_a$, $T$ is randomly sampled from a uniform distribution from 1 to 1000. The $\prod$ symbol denotes the projection function, and $\mathbb{X}_a = \{\boldsymbol{x}: \| \boldsymbol{x}-\boldsymbol{x}_s \| \leq \epsilon\ \&\ \boldsymbol{x} \in [0,1]^{H\times W \times 3} \}$. Figure~\ref{fig:fpatch} provides examples of F-patch.

\begin{figure*}[]
    \centering
    \subfigure[$\boldsymbol{M}$]{
		\includegraphics[width=0.2\linewidth]{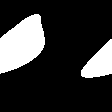}}
    \subfigure[$\nabla _{\boldsymbol{M}}$]{
		\includegraphics[width=0.2\linewidth]{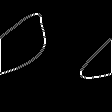}}
    \caption{An example of $\nabla _{\boldsymbol{M}}$.}
    \label{fig:grad_mask}
\end{figure*}
\subsection{Step 3: Train the Patch Segmenter}
\label{ebce}
 In this step, we train the patch segmenter with the F-patch using the EBCE loss function. We use U-Net~\cite{ronneberger2015u} as the network architecture for the patch segmenter. Let $\mathcal{F}_{PS}$ denote the patch segmenter and $\theta$ denote the parameters of the network. $\boldsymbol{x}_a$ denotes an attack image with F-patch and $\boldsymbol{M}$ is its ground-truth binary patch mask. $\mathcal{L}_{EBCE}$ denotes the EBCE loss function. We treat the challenge of detecting adversarial patches as a segmentation problem, and our solution involves training a dedicated patch segmenter. This training process can be formulated as an optimization problem:
\begin{equation}
\min_{\theta }\mathcal{L}_{EBCE}(\mathcal{F}_{PS}(\boldsymbol{x}_{a} ;\theta), \boldsymbol{M}),
\end{equation}
where $\mathcal{F}_{PS}(\boldsymbol{x}_{a} ;\theta) \in [0,1]^{H \times W}$ is the output probability map. To enhance the accuracy of detecting patch edges, we propose the EBCE loss function, which is formulated as follows:
\begin{equation}
\begin{aligned}
\mathcal{L}_{EBCE}(\hat{\boldsymbol{M}} ,\boldsymbol{M}) =-\frac{1}{HW}\sum _{i}^{H}\sum _{j}^{W} e^{\beta (\nabla _{\boldsymbol{M}})_{ij}}[&\boldsymbol{M}_{ij} \log\hat{\boldsymbol{M}}_{ij}\\&+( 1-\boldsymbol{M}_{ij}) \log( 1-\hat{\boldsymbol{M}}_{ij})], 
\end{aligned}
\end{equation}
where $\nabla _{\boldsymbol{M}}$ denotes the gradient of $\boldsymbol{M}$. Fig~\ref{fig:grad_mask} shows an example of $\nabla _{\boldsymbol{M}}$. The horizontal and vertical Sobel convolution kernels~\cite{sobel19683x3} are represented by $\boldsymbol{S}_{x}$ and $\boldsymbol{S}_{y}$, respectively. The gradient magnitude $\nabla _{\boldsymbol{M}}$ is computed as follows:
\begin{equation}
    \begin{aligned}
\nabla _{\boldsymbol{M}} & =\sqrt{(\boldsymbol{M} \otimes \boldsymbol{S}_{x})^{2} +(\boldsymbol{M} \otimes \boldsymbol{S}_{y})^{2}},\\
\boldsymbol{S}_{x} & = \begin{pmatrix}
1 & 0 & -1\\
2 & 0 & -2\\
1 & 0 & -1
\end{pmatrix},
\boldsymbol{S}_{y} = \begin{pmatrix}
1 & 2 & 1\\
0 & 0 & 0\\
-1 & -2 & -1
\end{pmatrix},
\end{aligned}
\end{equation}
where $\otimes$ denotes the convolutional product operator. After training the patch segmenter $\mathcal{F}_{PS}$, we can generate the initial binary patch mask, denoted as $\boldsymbol{M}_{PS}$, by binarizing the output probability map as follows:
\begin{equation}
    \boldsymbol{M}_{PS} = \mathcal{F}_{FR}(\boldsymbol{x}_a)>0.5.
\end{equation}

\begin{figure*}[]
    \centering
    \includegraphics[width=0.9\linewidth]{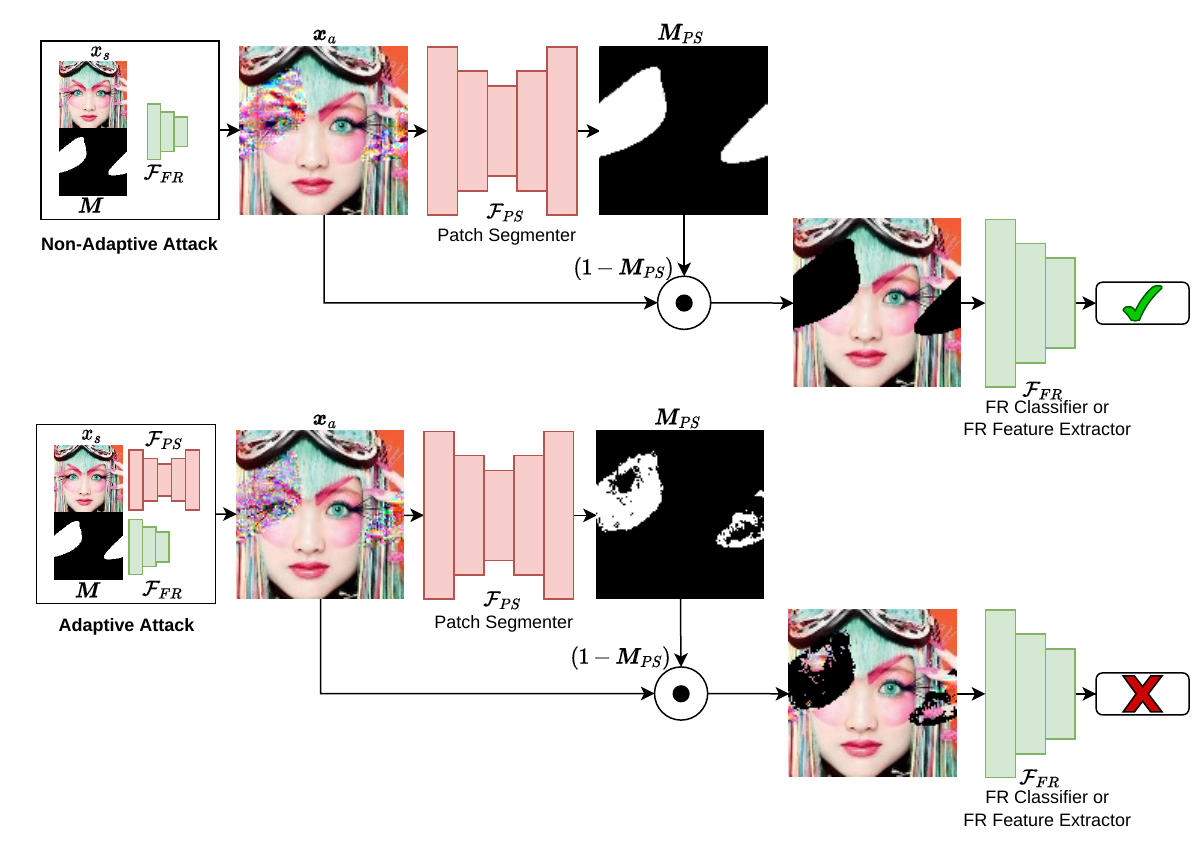}
    \caption{Examples of non-adaptive and adaptive attacks.}
    \label{fig:AA}
\end{figure*}

Figure~\ref{fig:AA} presents instances of both non-adaptive and adaptive attacks. In a non-adaptive attack scenario, the attacker solely relies on the FR model, and our patch segmenter efficiently identifies the adversarial patches, leading to a successful defense. However, it is essential to note that the patch segmenter $\mathcal{F}_{PS}$ also employs a neural network architecture, making it susceptible to adversarial examples. This vulnerability becomes particularly pronounced in the context of strong white-box adaptive attacks where the attacker possesses comprehensive knowledge of the system. In this paper, we employ the BPDA adaptive attack~\cite{athalye2018obfuscated}, which involves approximating the non-differentiable part of the defense system with a differentiable approximation to execute an effective adaptive attack. To achieve this, we use Sigmoid functions, denoted as $\sigma$, to approximate the non-differentiable binarization operation, as shown in the equation below:
\begin{equation}
  \sigma(\mathcal{F}_{FR}(\boldsymbol{x}_a)) \approx (\mathcal{F}_{FR}(\boldsymbol{x}_a)>0.5).
\end{equation}
As demonstrated in Figure~\ref{fig:AA}, the adaptive attack scenario exposes a critical vulnerability in our defense system when relying solely on a single patch segmenter. To address this challenge effectively, we introduce the SAF strategy, which is explained in Section~\ref{saf}.  

\begin{figure*}[]
    \centering
    \includegraphics[width=0.9\linewidth]{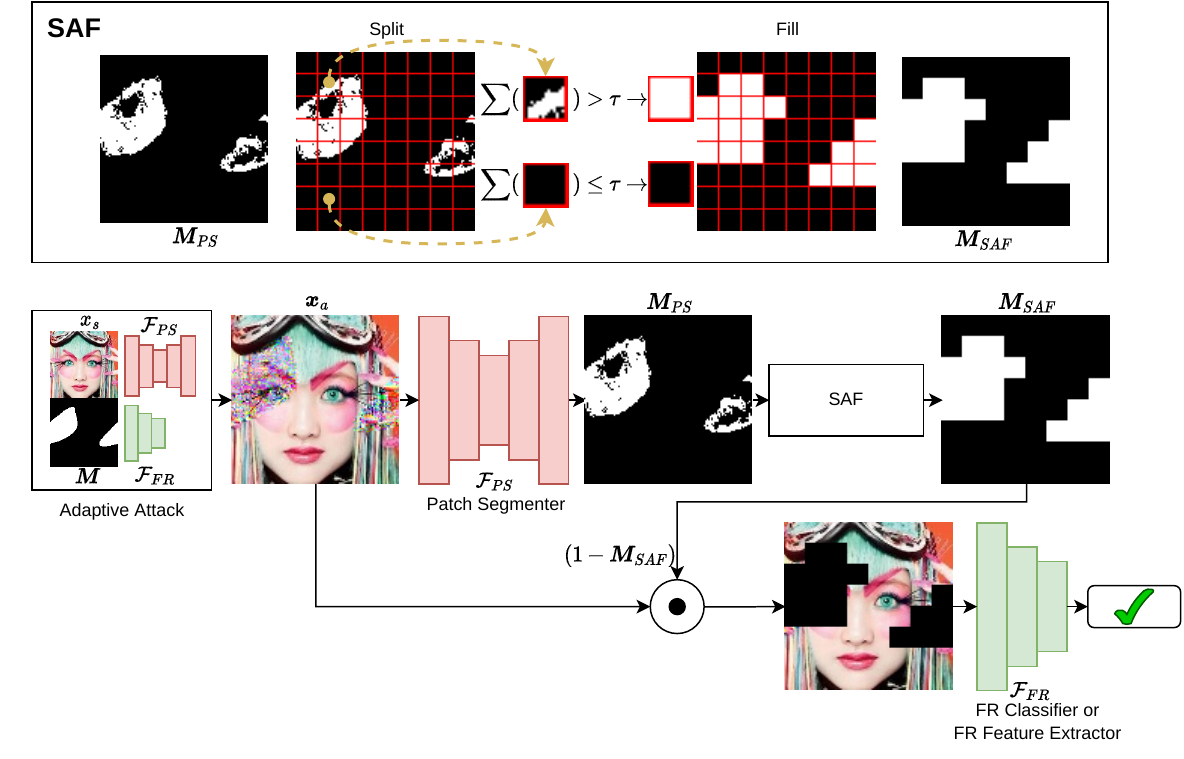}
    \caption{Split and Fill Strategy. $\tau$ denotes a predefined threshold.}
    \label{fig:saf}
\end{figure*}
\subsection{Step 4: Using the SAF Strategy}
\label{saf}
In this step, we introduce the SAF strategy to handle the initial patch mask. As depicted in Figure~\ref{fig:saf}, the SAF strategy serves as an effective defense against strong adaptive attacks, even when faced with scenarios involving potential patch segmenter information leakage.  

The SAF strategy entails several key steps: Initially, the initial patch mask is divided into smaller subgrids. Next, the sum of values within each subgrid is calculated. Subgrids with a sum exceeding a predefined threshold ($\tau$) are populated with 1s, while those below the threshold are filled with 0s. This process is applied iteratively to all subgrids, resulting in the final updated mask.

The adaptability of the SAF strategy to various adversarial patch shapes and quantities is evident in Figure~\ref{fig:saf}, highlighting its robustness against adaptive attacks. A comprehensive understanding of the SAF algorithm can be found in Algorithm~\ref{alg:saf}. It is noteworthy that the overall time complexity of the SAF algorithm is $\mathcal{O}(HW)$.
\begin{algorithm}
\caption{Split and Fill algorithm}
\begin{algorithmic}[1]
\REQUIRE Initial patch mask $\boldsymbol{M} \in [0,1]^{H \times W}$, with both the height and width ($H$ and $W$) being equal.
\REQUIRE Number of subgrids $n$.
\STATE Extract dimensions: $H$ and $W$ from $\boldsymbol{M}$.
\STATE Calculate subgrid side length: $g_{size} = \lfloor \frac{H}{n} \rfloor$.
\STATE Set the predefined threshold: $\tau = g_{size}$.
\FOR{$i$ \textbf{in} $[0, n)$}
    \FOR{$j$ \textbf{in} $[0, n)$}
        \STATE Compute subgrid boundaries:
        \STATE \hspace{\algorithmicindent} $g_h = \min((i + 1) \cdot g_{size}, H)$,
        \STATE \hspace{\algorithmicindent} $g_w = \min((j + 1) \cdot g_{size}, W)$.
        \STATE Calculate subgrid sum: 
        \STATE \hspace{\algorithmicindent} $g_{sum} = \sum_{x=i \cdot g_{size}}^{g_w} \sum_{y=j \cdot g_{size}}^{g_w} \boldsymbol{M}_{xy}$.
        \STATE Update subgrid:
        \IF {$g_{sum} > \tau$}
            \STATE Fill all elements of $\boldsymbol{M}[i \cdot g_{size}:g_h, j \cdot g_{size}:g_w]$ with 1. 
        \ELSE
            \STATE Fill all elements of $\boldsymbol{M}[i \cdot g_{size}:g_h, j \cdot g_{size}:g_w]$ with 0. 
        \ENDIF
    \ENDFOR
\ENDFOR
\RETURN updated mask $\boldsymbol{M}$
\end{algorithmic}
\label{alg:saf}
\end{algorithm}

\section{Experiments}
\label{sec:experiments}
In this section, we comprehensively evaluate the performance of RADAP in defending against a variety of adversarial patches on FR systems. Additionally, we assess the robustness of RADAP against adaptive attacks, especially in the case where the patch segmenter model is leaked. The experiments conducted aim to provide a thorough understanding of the effectiveness and capabilities of RADAP in FR systems.

\begin{figure*}[]
    \centering
    \includegraphics[width=\linewidth]{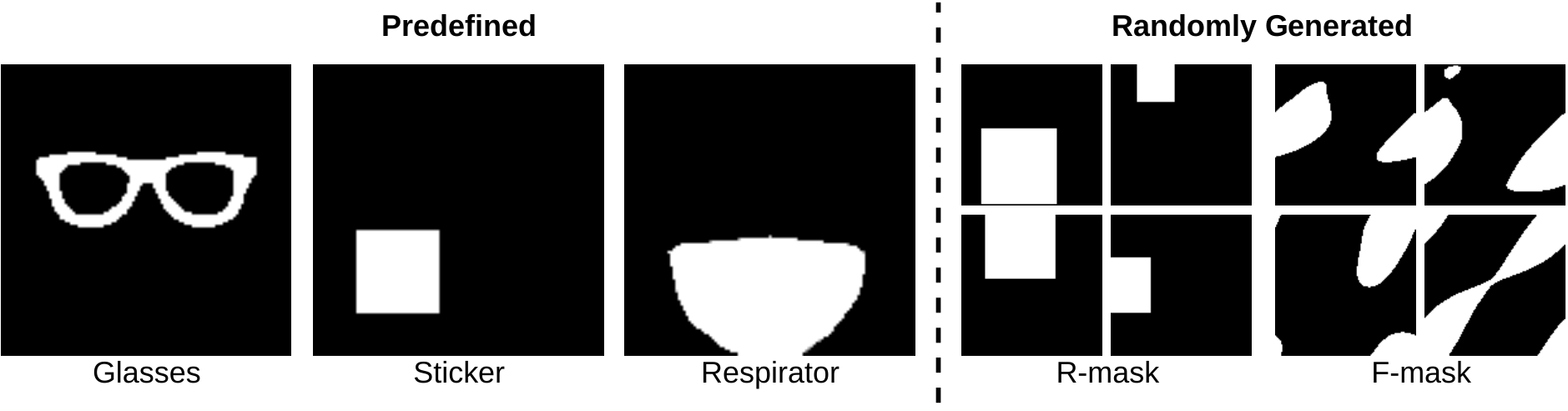}
    \caption{Examples of the patch masks for testing.}
    \label{fig:tpm}
\end{figure*}
\subsection{Experimental Setup}
\textbf{Datasets.} For closed-set FR systems, we use a subset of the VGGFace2 dataset~\cite{cao2018vggface2}. The subset contains 100 classes, and we split the images into training, validation, and testing sets in a 7:1:2 ratio. For open-set FR systems, we use the MS-Celeb-1M dataset~\cite{guo2016ms} for training and the LFW dataset~\cite{huang2008labeled} for testing. Face images from all datasets are aligned to a size of $112\times112$. 

\textbf{Attack Patch Masks.} Figure~\ref{fig:tpm} illustrates the patch masks used in our testing. These masks fall into two categories: a predefined set containing masks such as glasses, stickers, and a respirator mask, and a second category of randomly generated masks, referred to as the Random Rectangle Mask (R-mask) and the Fourier space sampling-based mask (F-mask).

\begin{table}[]
\centering
\caption{Target models of open-set FR systems. }
\scalebox{0.9}{
\begin{tabular}{@{}l|l|l@{}}
\toprule
Target Model & Neural Architecture & Loss       \\ \midrule
ArcMob       &MobileFaceNet~\cite{chen2018mobilefacenets}       & ArcFace~\cite{deng2019arcface}     \\
MagMob       &MobileFaceNet~\cite{chen2018mobilefacenets}        & MagFace~\cite{meng2021magface}     \\
MvMob        &MobileFaceNet~\cite{chen2018mobilefacenets}        & MV Softmax~\cite{wang2020mis}  \\
MvRes50      & IResNet50~\cite{he2016deep}            & MV Softmax~\cite{wang2020mis}  \\
MvSwinT      & Swin-T~\cite{liu2021swin}               & MV Softmax~\cite{wang2020mis}  \\ \bottomrule
\end{tabular}
}
\label{tab:Target_Model}
\end{table}

\textbf{Target Models.}  For closed-set FR systems, we use four well-known target models, including MobileFaceNet~\cite{chen2018mobilefacenets}, IResNet50~\cite{he2016deep}, EfficientNet-B0~\cite{tan2019efficientnet}, and Swin-T~\cite{liu2021swin}. For open-set FR systems, we employ five target models, summarized in Table~\ref{tab:Target_Model}.

\textbf{Attack Methods.}  In this experiment, we employ four distinct adversarial patch attack methods to evaluate the effectiveness of our defense approach. These methods consist of PGDAP~\cite{madry2017towards}, APGDAP~\cite{croce2020reliable}, TAP~\cite{xiao2021improving}, and RSTAM~\cite{liu2022rstam}. Specifically, PGDAP and APGDAP are the adversarial patch versions of the PGD~\cite{madry2017towards} and Auto-PGD~\cite{croce2020reliable} attacks. TAP and RSTAM leverage input diversity to enhance the transferability of adversarial patches, endowing them with more robust attack capabilities compared to PGDAP.

\textbf{Baseline Defense Methods.} We report the results of our method with SAF, \textbf{Ours($+$)}, and without SAF, \textbf{Ours($-$)}. Additionally, we compare our approach with five baseline adversarial patch defense methods: Defense against Occlusion Attacks (DOA)~\cite{wu2020defending}, JPEG Compression (JPEG)~\cite{dziugaite2016study}, Local Gradients Smoothing (LGS)~\cite{naseer2019local}, PatchCleanser (PC)~\cite{xiang2022patchcleanser}, and Segment and Complete (SAC)~\cite{liu2022segment}.

\textbf{Evaluation Metric.} For closed-set FR systems, our primary performance measure is classification accuracy (Acc). For open-set FR systems, we evaluate using True Acceptance Rate at a False Acceptance Rate of 0.001 (TAR@0.001FAR).

\textbf{Implementation Details.} During the training of the robust FR model, we set the values of $(a, b)$ as $(0, 1)$. When generating the F-patch, we configure the parameters $(a, b)$ to be $(0.02, 0.3)$. The patch segmenter is trained using the Stochastic Gradient Descent (SGD) optimizer~\cite{bottou2010large} with an initial learning rate of $0.1$, weight decay of $10^{-4}$, momentum of $0.9$, and Nesterov momentum enabled. We set the weight $\beta$ in EBCE to $1$. Moreover, we apply a cosine annealing scheduler technique~\cite{loshchilov2016sgdr} to adjust the learning rate schedule. For the attack,  the default attack method is PGDAP, the default number of perturbation iterations is set to $100$ steps, using a perturbation step size $\alpha$ of $0.007$, and the perturbation boundary $\epsilon$ is defined as $0.3$. The default value for the number of subgrids $n$ in the SAF strategy is set to $8$. 

Specifically, the attack method for generating the F-patch used to train the patch segmenter exclusively employs PGDAP, with the attacker's goal being evasion. Moreover, the target FR model is MobileFaceNet, which is trained on a closed-set dataset.

\begin{table}[]
\centering
\caption{Comparison results of different methods for defending unseen patch masks (shapes) and attacker's goals on closed-set FR systems. The best performances in each block are shown in \textbf{bold}. The evaluation metric uses Acc (\%).}
\scalebox{0.8}{
\begin{tabular}{@{}l|l|c|ccccc@{}}
\toprule
\multirow{2}{*}{\begin{tabular}[c]{@{}l@{}}Attacker’s \\ Goal\end{tabular}}        & \multirow{2}{*}{\begin{tabular}[c]{@{}l@{}}Defense\\ Method\end{tabular}}&\multirow{2}{*}{Clean} & \multicolumn{5}{c}{Mask}                         \\
                               &                         &         & Glasses & Sticker & Respirator & R-mask & F-mask \\ \midrule
\multirow{7}{*}{Evasion}     
                               & DOA~\cite{wu2020defending}              &80.07    &3.15     &24.58    &1.20        &30.77   & 12.47   \\
                               & JPEG~\cite{dziugaite2016study}            &88.57    &29.96    &32.44    &0.02        & 39.90  &14.19    \\
                               & LGS~\cite{naseer2019local}             &87.59    &47.26    &60.75    &13.35       & 54.90  &22.40    \\
                               & PC~\cite{xiang2022patchcleanser}              &92.70    &5.18     &63.21    &0.44        & 49.30  &12.10    \\
                               & SAC~\cite{liu2022segment}             &92.97    &30.76    &90.66    &79.24       & 88.98  &21.83    \\ \cmidrule(l){2-8} 
                               & \textbf{Ours($-$)}                  &94.00 &\textbf{91.39} & \textbf{92.55} & \textbf{87.46} & \textbf{92.11} & \textbf{78.86}  \\
                               & \textbf{Ours($+$)}      &\textbf{94.01}&84.72&91.28&83.55&89.97& 69.79    \\
                               \midrule
\multirow{7}{*}{Impersonation} 
                               & DOA~\cite{wu2020defending}              &   -     &29.21    &63.73    &16.35       &50.15   &22.59   \\
                               & JPEG~\cite{dziugaite2016study}            &   -     &57.00    &56.74    &11.43       & 52.15  &24.39   \\
                               & LGS~\cite{naseer2019local}             &   -     &68.04    &77.74    &43.72       & 72.24  &39.44    \\
                               & PC~\cite{xiang2022patchcleanser}              &   -     &30.69   &68.10    &20.55       & 63.52  &17.85   \\
                               & SAC~\cite{liu2022segment}             &   -     &61.99    &90.82    &80.14       & 89.53  &30.50    \\ \cmidrule(l){2-8} 
                               & \textbf{Ours($-$)}                  &   -     &\textbf{91.42} &\textbf{92.55} &\textbf{87.46}  & \textbf{92.23}  &\textbf{78.79}   \\
                               & \textbf{Ours($+$)} &  -    &84.89&91.28&84.23&90.24&70.56\\\bottomrule
\end{tabular}
}
\label{tab:1}
\end{table}

\subsection{Comparison Studies on Closed-set FR Systems}
\textbf{Defense against Unseen Patch Masks.} Table~\ref{tab:1} presents a comparative analysis of various defense methods against unseen patch masks in closed-set FR systems. In practice, attackers often use undisclosed patch shapes, making it crucial to establish robust defenses against a variety of shapes. Our proposed method consistently demonstrates strong defense capabilities against different patch masks, whether they are predetermined or randomly generated. Notably, it outperforms other methods, particularly in scenarios involving Glasses masks and F-masks. For Glasses masks, ``\textbf{Ours($-$)}" outperforms ``SAC'' by 60.63\% (91.39\% vs. 30.76\%), and ``\textbf{Ours($+$)}" surpasses ``SAC'' by 53.96\% (84.72\% vs. 30.76\%). Additionally, our method achieves the highest performance in clean classification accuracy, indicating minimal information loss when defending against clean data.

\begin{figure*}[!ht]
    \centering
    \subfigure[Glasses]{
		\includegraphics[width=0.35\linewidth]{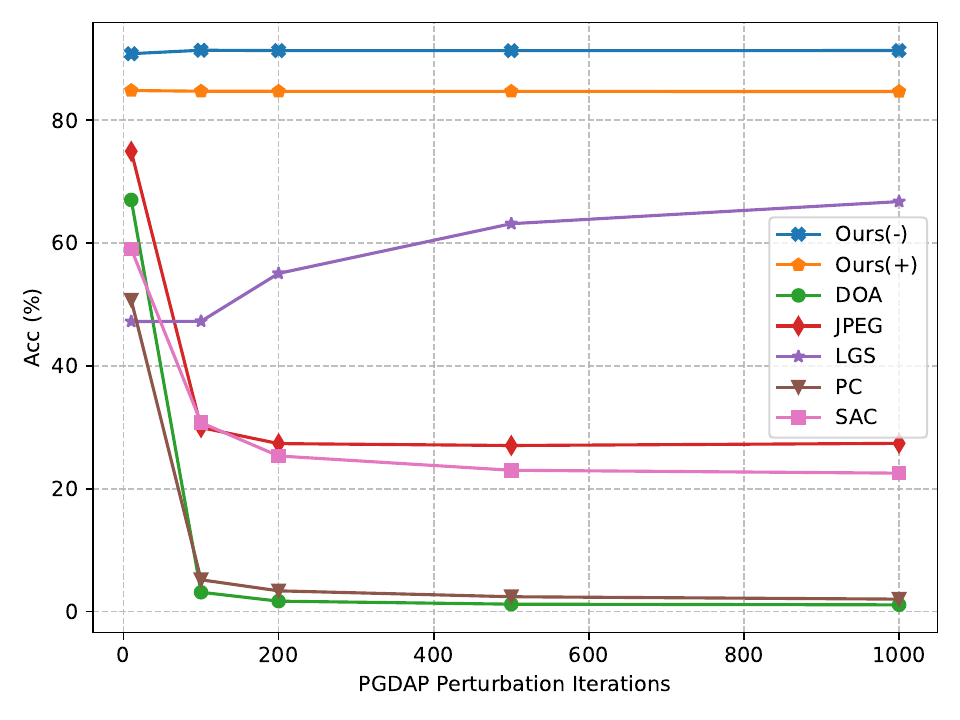}}
    \subfigure[Sticker]{
		\includegraphics[width=0.35\linewidth]{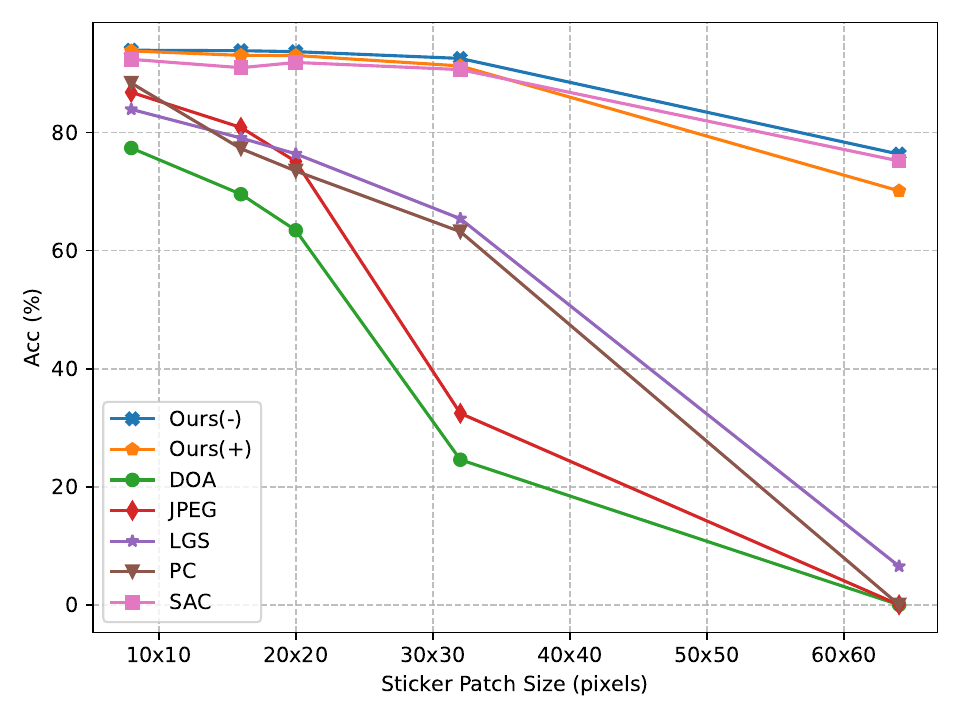}}
    \subfigure[R-mask]{
		\includegraphics[width=0.35\linewidth]{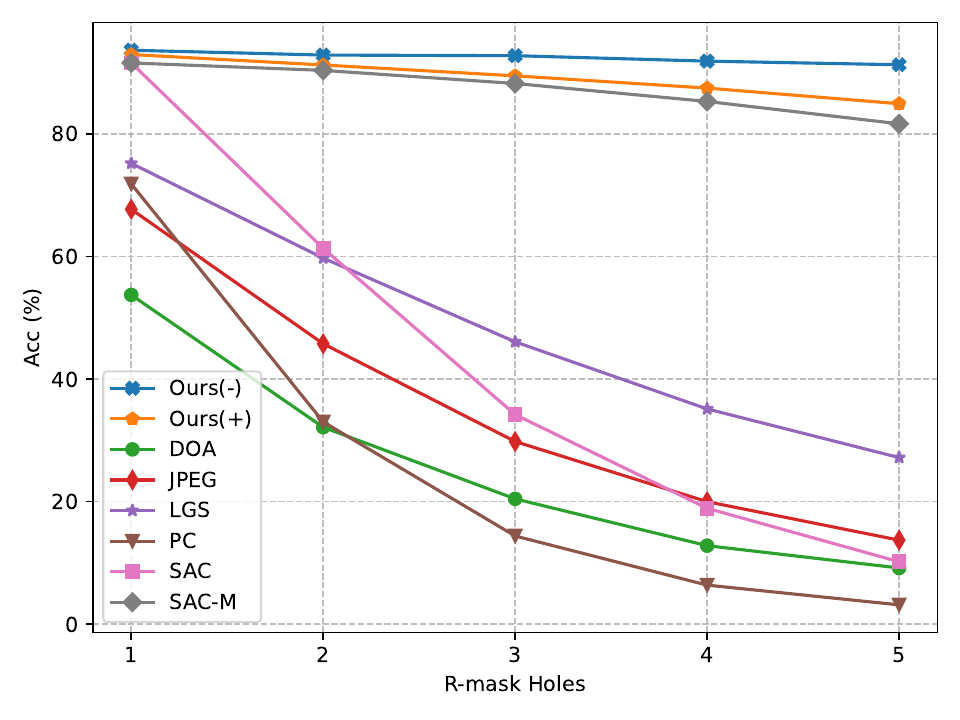}}
  \subfigure[F-mask]{
		\includegraphics[width=0.35\linewidth]{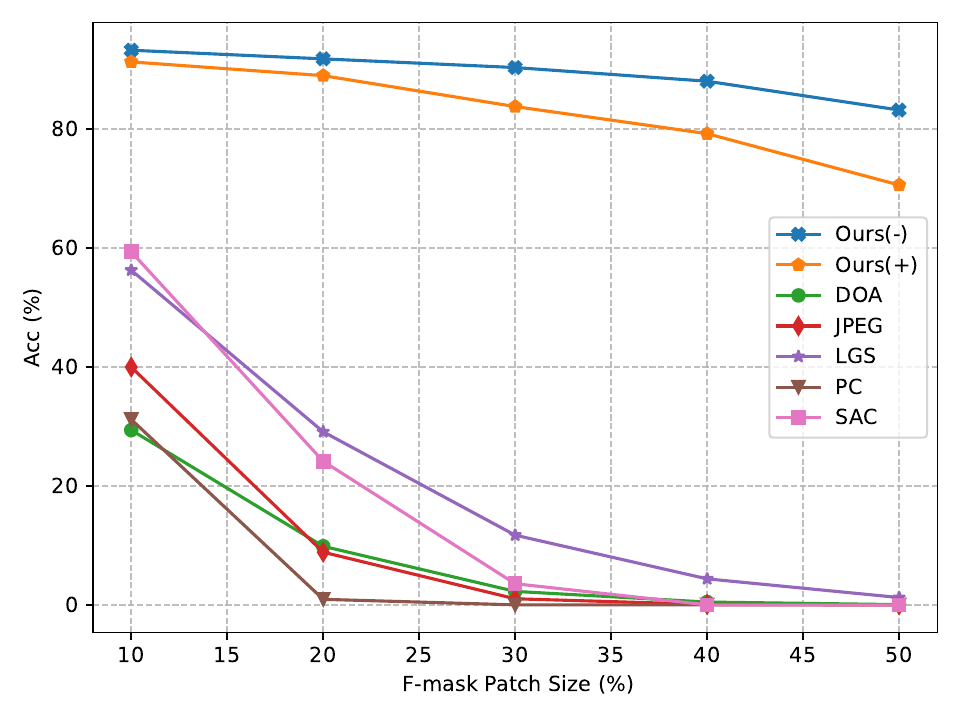}}
    \caption{Comparison results under different parameters for different unseen patch masks.}
    \label{fig:patch_mask}
\end{figure*}

Figure~\ref{fig:patch_mask} provides a comprehensive comparison of defense methods across various unseen patch masks with different parameters. Our method consistently demonstrates robust defense performance across different unseen patch masks and parameters. Notably, For R-masks, where SAC-M denotes the SAC version that specifies the number of holes. However, in reality, the number of holes employed by an attacker is often unknown to us. Our method's adaptability eliminates the need to specify the number of holes, ensuring effective defense regardless of the attacker's choices.

\textbf{Defense against Unseen Attacker's Goals.} Table~\ref{tab:1} also presents results when the attacker's goal is impersonation. Although our patch segmenter was not specifically trained for impersonation attacks, our method exhibits superior performance in these scenarios, showcasing its adaptability.

\begin{table}[]
\centering
\caption{Comparison results of different methods for defending unseen target models on closed-set FR systems.}
\scalebox{0.8}{
\begin{tabular}{@{}l|l|c|ccccc@{}}
\toprule
\multirow{2}{*}{Target Model}        & \multirow{2}{*}{\begin{tabular}[c]{@{}l@{}}Defense\\ Method\end{tabular}}&\multirow{2}{*}{Clean} & \multicolumn{5}{c}{Mask}                         \\
                               &                         &    & Glasses & Sticker & Respirator & R-mask & F-mask \\ \midrule
\multirow{6}{*}{IResNet50~\cite{he2016deep}}     
                               & JPEG~\cite{dziugaite2016study}            &90.75 & 24.47&32.27 &0.24  &40.93 & 14.55\\
                               & LGS~\cite{naseer2019local}             &91.46  & 55.04&  67.66 & 18.63 & 64.01&28.23  \\
                               & PC~\cite{xiang2022patchcleanser}              &94.52  &2.08  &73.89   &0.64    &56.88 &14.14 \\ 
                               & SAC~\cite{liu2022segment}             &95.12 &24.75& 93.59 & 87.25  &92.04  & 24.81 \\ \cmidrule(l){2-8}
                               & \textbf{Ours($-$)}         &\textbf{95.13} & \textbf{93.87} & \textbf{93.89}  & \textbf{89.95}  &\textbf{93.73} &\textbf{83.78} \\ 
                               & \textbf{Ours($+$)} &  94.99    &89.54&92.87&86.77&92.14&74.62\\\midrule
\multirow{6}{*}{EfficientNet-B0~\cite{chen2018mobilefacenets}}     
                               & JPEG~\cite{dziugaite2016study}            &90.80&34.80&  50.36 & 5.29 & 52.38 &  21.99\\
                               & LGS~\cite{naseer2019local}             &86.16 & 59.82 & 74.89 & 48.23 &  73.30 & 41.40 \\
                               & PC~\cite{xiang2022patchcleanser}              &93.83&0.37&68.04&0.41& 51.12&12.26\\
                               & SAC~\cite{liu2022segment}             &94.39&11.61&92.81&83.86&90.34& 21.81  \\ \cmidrule(l){2-8}
                               & \textbf{Ours($-$)}         &\textbf{94.48}& \textbf{92.27}& \textbf{92.96} &  \textbf{88.51}  &  \textbf{92.48}&\textbf{ 81.12} \\ 
                               & \textbf{Ours($+$)} &  94.22    &86.05&91.77&84.86&90.50&72.26\\\midrule
\multirow{6}{*}{Swin-T~\cite{liu2021swin}} 
                               & JPEG~\cite{dziugaite2016study}            &82.79& 43.54  &  61.53  & 6.44    & 51.11  &20.63 \\
                               & LGS~\cite{naseer2019local}             &89.89&42.95 &  65.51 &  10.15  & 54.55 &  21.34  \\
                               & PC~\cite{xiang2022patchcleanser}              & 92.14&0.43&73.50&0.00&37.68&10.26\\
                               & SAC~\cite{liu2022segment}             &92.38&18.94&91.28&82.16&89.04&20.99\\  \cmidrule(l){2-8}
                               & \textbf{Ours($-$)}         &\textbf{92.41} &\textbf{91.25} & \textbf{91.51 } & \textbf{87.06}  &\textbf{91.07}  & \textbf{80.88} \\
                               & \textbf{Ours($+$)} &  91.37    &87.01&90.10&81.60&89.25&72.66\\\bottomrule
\end{tabular}
}
\label{tab:2}
\end{table}

\textbf{Defense against Unseen Target Models.} In our approach, we utilize the MobileFaceNet as the target model for generating the F-patch used in patch segmenter training. In Table~\ref{tab:2}, we present a comparative analysis of our defense method's performance when confronted with previously unseen target models, including the transformer-based model Swin-T. Remarkably, our defense method consistently delivers strong performance across different target models.

\begin{table}[]
\centering
\caption{Comparison results of different methods for defending unseen attack methods on closed-set FR systems.}
\scalebox{0.8}{
\begin{tabular}{@{}l|l|c|ccccc@{}}
\toprule
\multirow{2}{*}{Attack Method}        & \multirow{2}{*}{\begin{tabular}[c]{@{}l@{}}Defense\\ Method\end{tabular}}&\multirow{2}{*}{Clean} & \multicolumn{5}{c}{Mask}                         \\
                               &                         &     & Glasses & Sticker & Respirator & R-mask & F-mask \\ \midrule
\multirow{6}{*}{APGDAP~\cite{croce2020reliable}} & DOA~\cite{wu2020defending}            &80.07&12.42&36.66&15.18&37.76&20.61 \\
                               & JPEG~\cite{dziugaite2016study}            &88.57&38.18&47.28&8.07& 46.10&19.28\\
                               & LGS~\cite{naseer2019local}             &87.59&71.46&79.70&52.34&74.86&42.29\\
                               & PC~\cite{xiang2022patchcleanser}              &92.70&28.95&66.92&9.06&54.51&19.60\\
                               & SAC~\cite{liu2022segment}             &92.76&42.82&91.13&79.14&89.38&29.04\\ \cmidrule(l){2-8}
                               & \textbf{Ours($-$)}         &{94.00}&\textbf{91.66}&\textbf{92.90}&\textbf{88.49}&\textbf{92.36}&\textbf{79.87}\\ 
                               & \textbf{Ours($+$)} &  \textbf{94.01}    &85.69&91.91&84.51& 90.73&71.12\\ \midrule
\multirow{6}{*}{TAP~\cite{xiao2021improving}}   & DOA~\cite{wu2020defending}              &80.14&4.00&31.24&2.36&33.16&13.94\\
                               & JPEG~\cite{dziugaite2016study}            &88.43&6.99&13.27&0.00&28.59&9.31\\
                               & LGS~\cite{naseer2019local}             &87.59&45.84&47.72&1.88& 43.22&15.88 \\
                               & PC~\cite{xiang2022patchcleanser}              &92.68&12.02& 63.72&0.67&50.88&13.74\\
                               & SAC~\cite{liu2022segment}             &92.12&42.40&90.30&78.93&89.08&24.77\\ \cmidrule(l){2-8}
                               & \textbf{Ours($-$)}         &{94.00}&\textbf{89.96}&\textbf{92.46}&\textbf{87.40}&\textbf{92.05}&\textbf{78.13}\\ 
                               & \textbf{Ours($+$)} &  \textbf{94.01}    &84.61&91.31&82.99&90.01&69.39\\ \midrule
\multirow{6}{*}{RSTAM~\cite{liu2022rstam}} & DOA~\cite{wu2020defending}              &80.14&13.34&51.13&13.59&42.55&19.28\\
                               & JPEG~\cite{dziugaite2016study}            &88.43&20.22&37.50&0.05&39.96&14.96\\
                               & LGS~\cite{naseer2019local}             &87.59&48.96&57.22&5.14&50.77&20.47\\
                               & PC~\cite{xiang2022patchcleanser}              &92.68&37.21&66.96&4.58&58.13&19.81\\
                               & SAC~\cite{liu2022segment}             &92.12&53.98&90.49&79.22&89.23&30.45  \\ \cmidrule(l){2-8}
                               & \textbf{Ours($-$)}         &{94.00}&\textbf{90.90}&\textbf{92.49}&\textbf{87.39}&\textbf{92.00}&\textbf{78.37}\\ 
                               & \textbf{Ours($+$)} & \textbf{94.01}    &84.68&91.31&83.18&90.04&69.82\\ \bottomrule
\end{tabular}
}
\label{tab:3}
\end{table}

\textbf{Defense against Unseen Attack Methods.}  Table~\ref{tab:3} presents the defense results when subjected to three attack methods stronger than PGDAP, specifically APGDAP, TAP, and RSTAM. The noteworthy performance of our method in defending against these advanced attacks underscores its remarkable robustness.

\begin{table}[]
\centering
\caption{Results of ablation studies with FCutout. UN means undefended and GT is defended using ground truth mask.}
\scalebox{0.8}{
\begin{tabular}{@{}ll|c|ccccc@{}}
\toprule
\multicolumn{2}{c|}{\multirow{2}{*}{Defense Method}}&\multirow{2}{*}{Clean}               & \multicolumn{5}{c}{Mask}                                 \\
\multicolumn{2}{c|}{}                                      &       & Glasses & Sticker & Respirator & R-mask & F-mask \\ \midrule
\multicolumn{1}{l|}{\multirow{3}{*}{UN}}         & Vanilla & 89.43 & 0.00   &  0.04       &  0.00  &  14.12  & 5.51       \\
\multicolumn{1}{l|}{}                            & Cutout  &92.97 &  0.00    & 0.05 &0.00      &  13.85 & 5.71  \\
\multicolumn{1}{l|}{}                            & FCutout &\textbf{ 94.09}  &0.00    &\textbf{0.32}  & 0.00           &\textbf{17.22} &\textbf{6.40} \\ \midrule
\multicolumn{1}{l|}{\multirow{3}{*}{GT}}         & Vanilla &89.43  &64.23   &82.96  &  61.95 & 79.27& 55.32  \\
\multicolumn{1}{l|}{}                            & Cutout  &92.97 &88.70 & 90.83 &83.08 &90.82 &71.93 \\
\multicolumn{1}{l|}{}                            & FCutout & \textbf{94.09} &\textbf{91.42}  &\textbf{92.56} &\textbf{87.46}  & \textbf{92.12} & \textbf{78.90}\\ \midrule
\multicolumn{1}{l|}{\multirow{3}{*}{SAC~\cite{liu2022segment}}}& Vanilla &89.31 &3.69  & 82.21 & 48.92      & 72.20  &17.58  \\
\multicolumn{1}{l|}{}                            & Cutout  &92.97    &\textbf{30.76}    &90.66    &79.24       & 88.98  &21.83        \\
\multicolumn{1}{l|}{}                            & FCutout &\textbf{94.02}  &29.50 & \textbf{92.49}  &\textbf{83.47 }     & \textbf{90.17}  &\textbf{ 24.09} \\ \midrule
\multicolumn{1}{l|}{\multirow{3}{*}{\textbf{Ours($-$)}}}       & Vanilla &89.43 &63.81 &  82.97  & 61.94 & 79.29 & 55.06  \\
\multicolumn{1}{l|}{}                            & Cutout  &92.97&88.62 & 90.85 &  83.08   &90.79  & 71.89 \\
\multicolumn{1}{l|}{}                            & FCutout &\textbf{94.00} &\textbf{91.39} & \textbf{92.55} & \textbf{87.46} & \textbf{92.11} & \textbf{78.86}         \\  \midrule
\multicolumn{1}{l|}{\multirow{3}{*}{\textbf{Ours($+$)}}}       & Vanilla &88.62 &48.09 &  74.96  & 51.38 & 73.51 & 42.64  \\
\multicolumn{1}{l|}{}                            & Cutout  &92.86&\textbf{85.14} & 89.60 &  79.78   &88.90  &  65.19 \\
\multicolumn{1}{l|}{}                            & FCutout &\textbf{94.01}&{84.72}&\textbf{91.28}&\textbf{83.55}&\textbf{89.97}& \textbf{69.79}         \\
\bottomrule
\end{tabular}
}
\label{tab:dod_cutout}
\end{table}
\subsection{Ablation Studies on Closed-set FR Systems}
In this subsection, we perform ablation studies on closed-set FR systems, focusing on three key components: FCutout, F-patch, and EBCE.

\textbf{Ablation Studies with FCutout.} We start by examining the results of ablation studies with FCutout, and the findings are presented in Table~\ref{tab:dod_cutout}. The table clearly indicates that FCutout outperforms both Cutout and Vanilla in terms of clean classification accuracy. This observation highlights FCutout's superior generalization capabilities when compared to Cutout. Notably, FCutout consistently achieves higher classification accuracy than Cutout for most masks, underscoring its exceptional occlusion robustness.

\begin{table}[]
\centering
\caption{Results of ablation studies with F-patch and EBCE.}
\scalebox{0.8}{
\begin{tabular}{@{}l|c|ccccc@{}}
\toprule
\multicolumn{1}{c|}{\multirow{2}{*}{Defense Method}}  &\multirow{2}{*}{Clean} & \multicolumn{5}{c}{Mask}        \\
                      &    & Glasses & Sticker & Respirator & R-mask & F-mask \\ \midrule
R-patch + BCE           &\textbf{94.08} &78.55&92.53&86.47&92.11&65.87\\
R-patch + EBCE          &{94.02} &\textbf{90.64}&\textbf{92.53}&\textbf{87.25}&\textbf{92.12}&\textbf{77.57}        \\\midrule
F-patch + BCE           &93.93         &91.20&92.30&87.34&92.05&78.76\\
F-patch + EBCE          &\textbf{94.00} &\textbf{91.39} & \textbf{92.55} & \textbf{87.46} & \textbf{92.11} & \textbf{78.86}        \\
\bottomrule
\end{tabular}
}
\label{tab:fpatch_and_ebce}
\end{table}


\begin{figure*}[!ht]
    \centering
    \includegraphics[width=0.8\linewidth]{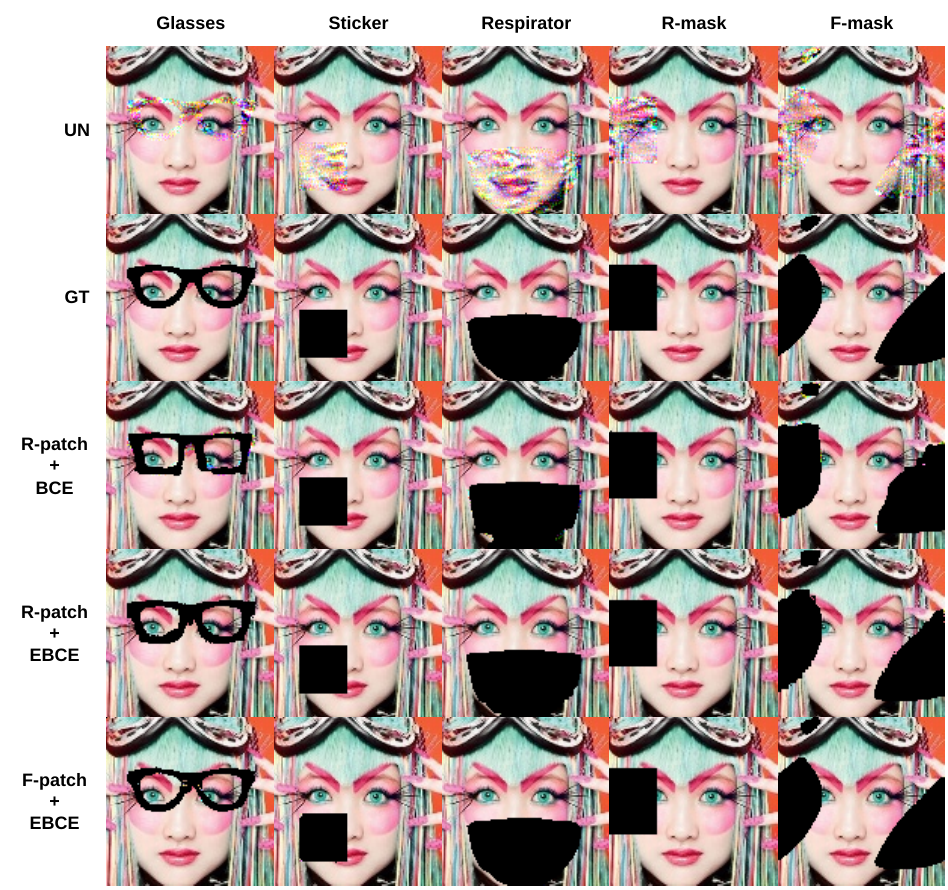}
    \caption{Visualization results of ablation studies with F-patch and EBCE.}
    \label{fig:fpatch_and_ebce}
\end{figure*}
\textbf{Ablation Studies with F-patch and EBCE.} Next, we delve into the results of ablation studies involving F-patch and EBCE, as detailed in Table~\ref{tab:fpatch_and_ebce}. In these studies, EBCE stands out by surpassing the performance of BCE, demonstrating its superior capabilities. Additionally, when comparing F-patch to R-patch, we find that F-patch generates more diverse patches, which significantly contributes to the improved generalization capacity of our defense method. For a more visual representation of these findings, we provide visualization results in Figure~\ref{fig:fpatch_and_ebce}, clearly illustrating the effectiveness of ``R-patch+EBCE'' in managing patch edges compared to ``R-patch+BCE''. it is worth noting that the combination of ``F-patch'' and ``EBCE'' yields the highest performance.

These ablation studies offer valuable insights into the individual contributions of FCutout, F-patch, and EBCE to the overall defense mechanism, confirming their significance in enhancing the robustness of our system.

\subsection{Generalization to Open-set FR Systems}
\begin{table}[]
\centering
\caption{Evaluation of defense results in the context of generalization to open-set FR systems, utilizing the TAR@0.001FAR (\%) evaluation metric.}
\scalebox{0.75}{
\begin{tabular}{@{}l|l|c|ccccc@{}}
\toprule
\multirow{2}{*}{Target Model}        & \multirow{2}{*}{\begin{tabular}[c]{@{}l@{}}Defense\\ Method\end{tabular}}&\multirow{2}{*}{Clean} & \multicolumn{5}{c}{Mask}                         \\
                               &                         &         & Glasses & Sticker & Respirator & R-mask & F-mask \\ \midrule
\multirow{4}{*}{ArcMob}        & JPEG~\cite{dziugaite2016study}            &91.80&8.36&42.30&0.10&36.63&22.43\\
                               & LGS~\cite{naseer2019local}             &88.80&32.83&89.36&50.40&83.86&79.50\\
                               & SAC~\cite{liu2022segment}             &98.16&0.00&96.96&37.83&82.63&0.00\\ \cmidrule(l){2-8}
                               & \textbf{Ours($-$)}                  &\textbf{98.20}&\textbf{83.93}&{97.06}&{66.30}&\textbf{92.40}&\textbf{90.16}\\ 
                               & \textbf{Ours($+$)}    &98.03 &73.10&\textbf{97.10}&\textbf{66.60}&89.23&86.96\\\midrule
\multirow{4}{*}{MagMob}        & JPEG~\cite{dziugaite2016study}            &89.63&8.70&43.80&0.23&34.03&25.36\\
                               & LGS~\cite{naseer2019local}             &91.33&29.43&90.53&57.03&83.50&79.90 \\ 
                               & SAC~\cite{liu2022segment}             &\textbf{98.33}& 0.66&97.50&56.06&85.93& 0.00\\ \cmidrule(l){2-8}
                               & \textbf{Ours($-$)}                  &98.30&\textbf{79.63}&\textbf{97.83}&\textbf{72.10}&\textbf{91.43}&\textbf{89.36}\\ 
                               & \textbf{Ours($+$)} & 98.10    &76.03&97.60&66.90&89.30&87.20\\\midrule
\multirow{4}{*}{MvMob}        & JPEG~\cite{dziugaite2016study}             &90.80&6.56&38.40&0.13&34.20&25.36\\
                               & LGS~\cite{naseer2019local}             &91.13&42.16&87.66&57.53&83.53&81.40 \\
                               & SAC~\cite{liu2022segment}             &\textbf{98.33}&0.33&\textbf{97.23}& 56.33&86.30&0.00   \\ \cmidrule(l){2-8}
                               & \textbf{Ours($-$)}                    &98.30&\textbf{85.13}& 97.06&\textbf{70.10}&\textbf{92.80}&\textbf{89.86}\\  
                               & \textbf{Ours($+$)} &  98.03   & 84.13&97.16&68.30&91.70&88.20\\\midrule
\multirow{4}{*}{MvRes50}        & JPEG~\cite{dziugaite2016study}           &97.90&49.06& 74.10& 1.96&52.70&44.26\\
                               & LGS~\cite{naseer2019local}             &97.63& 73.66&96.46&74.26&91.96& 91.96\\
                               & SAC~\cite{liu2022segment}             &\textbf{99.53}&6.46&99.23& 81.80& 94.60&0.20   \\ \cmidrule(l){2-8}
                               & \textbf{Ours($-$)}                  &\textbf{99.53}& \textbf{97.93}&\textbf{99.30}&\textbf{92.80}&\textbf{97.53}&\textbf{97.50}\\ 
                               & \textbf{Ours($+$)} & \textbf{99.53}    &96.46&99.23&90.60&96.50&96.10\\\midrule
\multirow{4}{*}{MvSwinT}       & JPEG~\cite{dziugaite2016study}            &98.20&91.00&94.46&76.73&90.86&91.96\\
                               & LGS~\cite{naseer2019local}             &99.30& 97.10&99.46&92.86& 96.96&97.03\\
                               & SAC~\cite{liu2022segment}             &99.60&2.23&\textbf{99.60}&85.76&94.60&0.00\\ \cmidrule(l){2-8}
                               & \textbf{Ours($-$)}                  &\textbf{99.66}&\textbf{98.83}&99.50&\textbf{95.56}&\textbf{97.60}& \textbf{97.96} \\
                               & \textbf{Ours($+$)} & 99.53    & 97.73&  99.46&93.20&96.70&96.36\\ \bottomrule
\end{tabular}
}
\label{tab:openset2}
\end{table}
 In contrast to closed-set FR systems, open-set FR systems find more frequent applications, particularly in commercial systems. In our approach, we opted not to retrain the patch segmenter but instead transferred it directly to open-set FR systems. Table~\ref{tab:openset2} provides an overview of the defense results when generalized to open-set FR systems. Once more, when confronted with unseen target models and unfamiliar patch shapes, our method exhibits superior defense performance. This observation underscores our approach's ability to maintain robust defense performance in open-set FR systems.

\subsection{Adaptive Attacks after Defense Method Leakage}

\begin{table}[]
\centering
\caption{Results of defense against adaptive attacks on closed-set FR systems. The attack method is PGDAP, the target model is MobileFaceNet, and the attacker's goal is evasion.}
\scalebox{0.9}{
\begin{tabular}{@{}l|c|ccccc@{}}
\toprule
 \multirow{2}{*}{\begin{tabular}[c]{@{}l@{}}Defense\\ Method\end{tabular}}&\multirow{2}{*}{Clean} & \multicolumn{5}{c}{Mask}                         \\
               &         & Glasses & Sticker & Respirator & R-mask & F-mask \\ \midrule
DOA~\cite{wu2020defending}     &80.07    &3.15     &24.58    &1.20        &30.77   & 12.47    \\
JPEG~\cite{dziugaite2016study}   &88.57    &1.33 &2.94&0.00  &21.48&7.29   \\
LGS~\cite{naseer2019local}    &87.59    &18.09&25.11&0.00&28.47&9.53 \\
PC~\cite{xiang2022patchcleanser}     &92.70    &5.18     &63.21    &0.44   & 49.30  &12.10        \\
SAC~\cite{liu2022segment}    &92.97    &14.42&88.88&69.83&85.37&26.29\\ \midrule
\textbf{Ours($-$)}&{94.00} &16.45&71.90&4.61&61.38&26.40\\ 
\textbf{Ours($+$)}&\textbf{94.01} &\textbf{82.85}&\textbf{91.15}&\textbf{81.81}&\textbf{89.76}&\textbf{69.75}\\ \bottomrule
\end{tabular}
}
\label{tab:bpda}
\end{table}

\begin{figure*}[]
    \centering
    \subfigure[Glasses]{
		\includegraphics[width=0.3\linewidth]{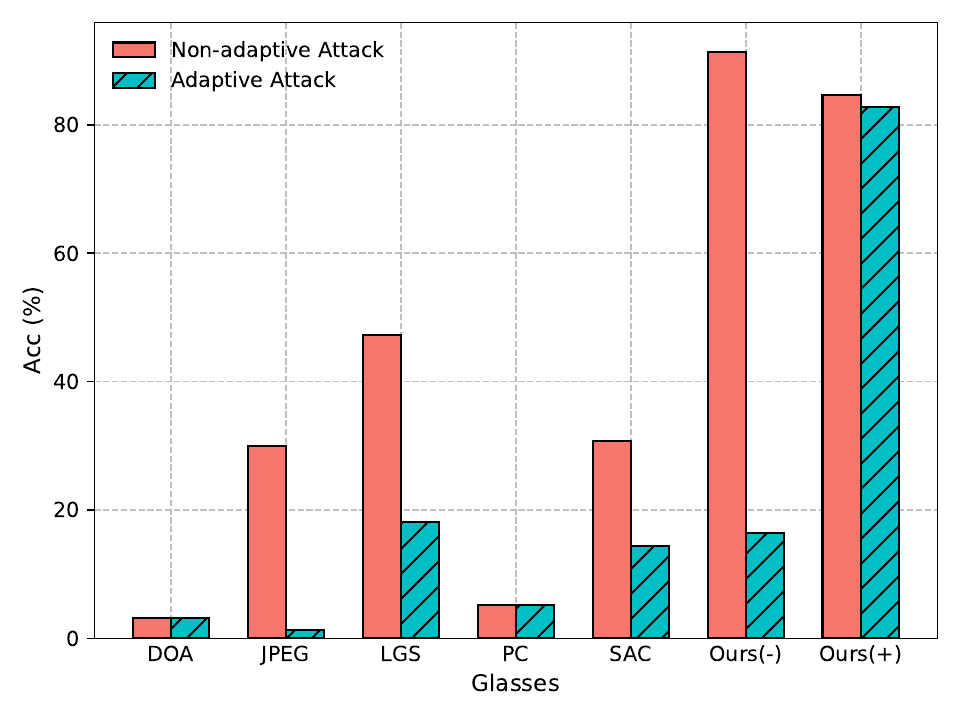}}
    \subfigure[Sticker]{
		\includegraphics[width=0.3\linewidth]{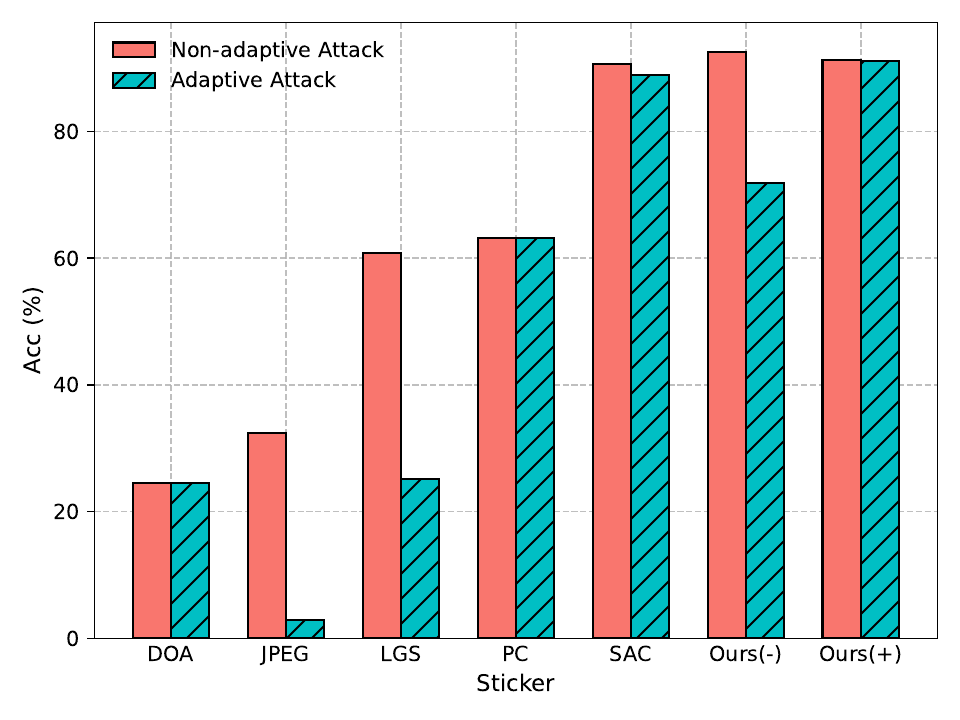}}
  \subfigure[Respirator]{
    		\includegraphics[width=0.3\linewidth]{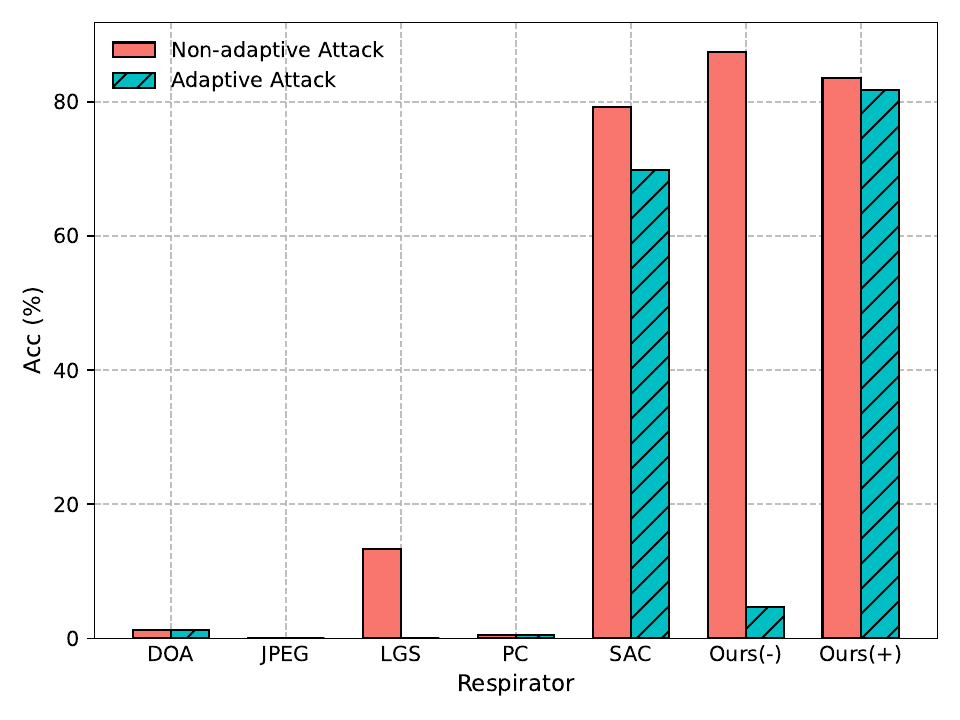}}
\subfigure[R-mask]{
		\includegraphics[width=0.3\linewidth]{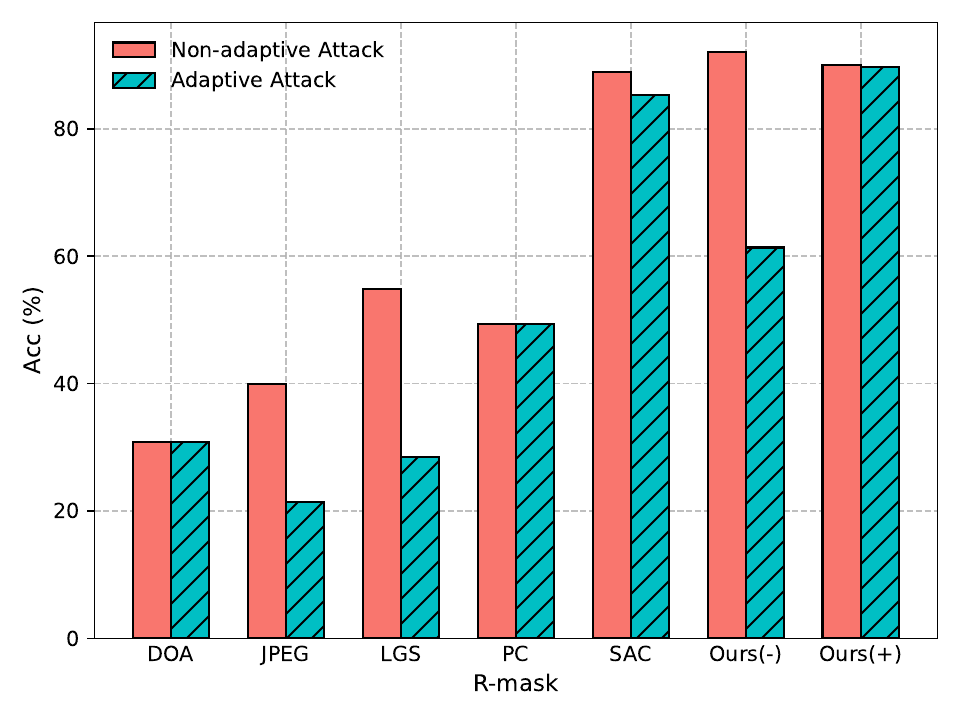}}
  \subfigure[F-mask]{
		\includegraphics[width=0.3\linewidth]{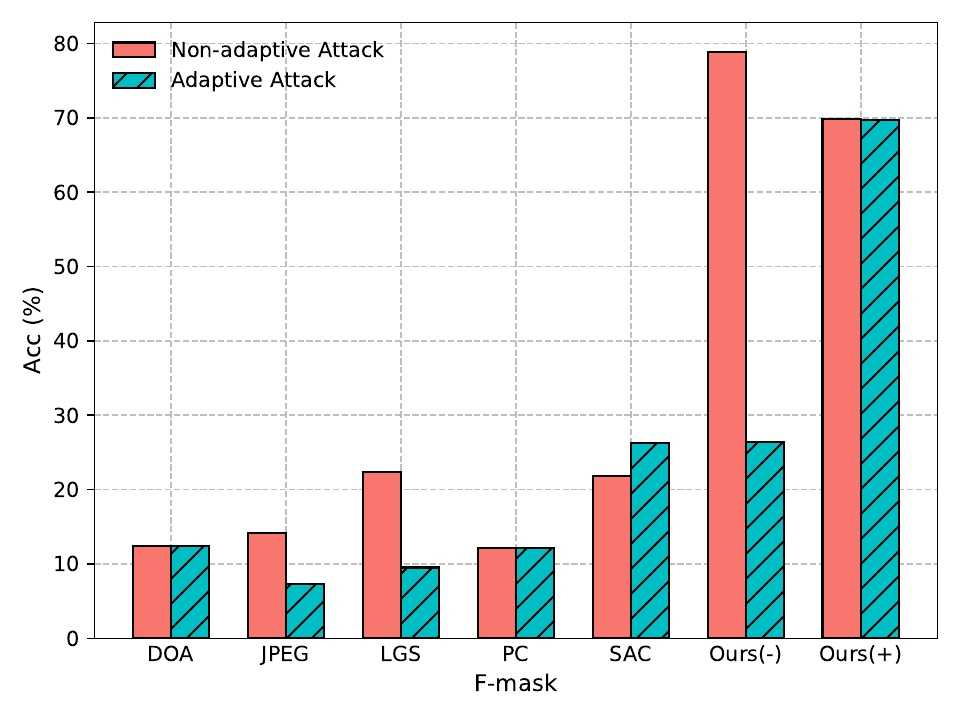}}
    \caption{Comparative results of non-adaptive and adaptive attacks.}
    \label{fig:non-a-c}
\end{figure*}

\begin{figure*}[!ht]
    \centering
    \includegraphics[width=0.9\linewidth]{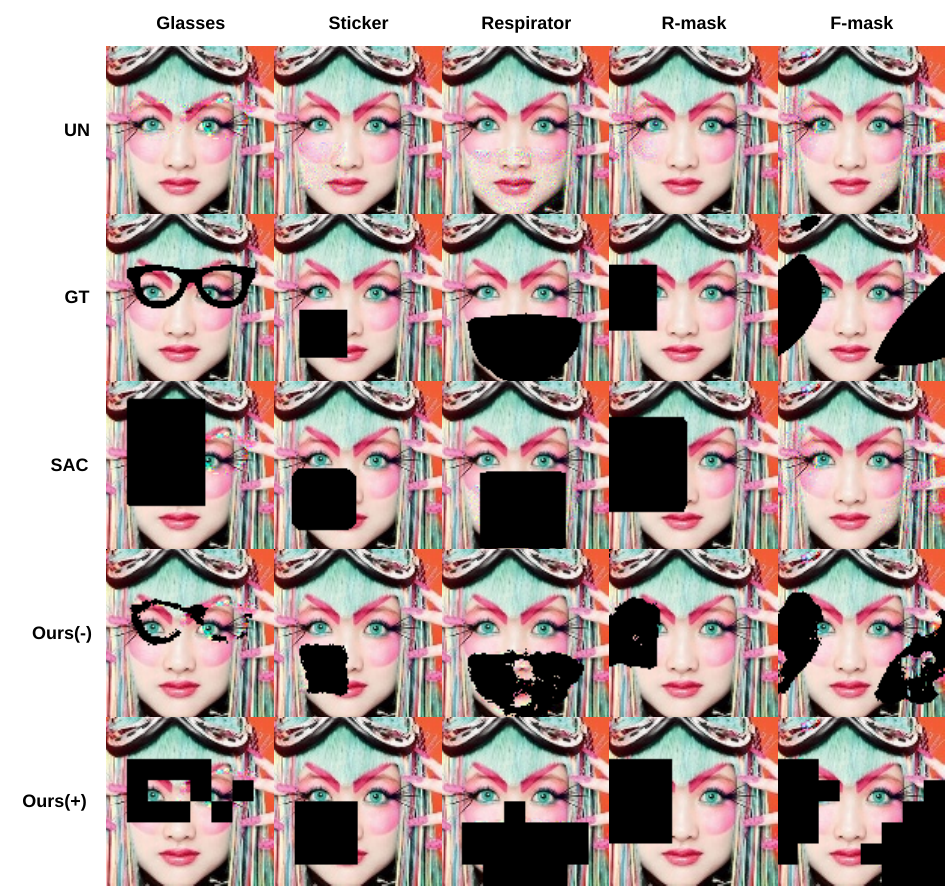}
    \caption{Visualization Results for Defending Against Adaptive Attacks.}
    \label{fig:A_attack}
\end{figure*}

\begin{figure*}[!ht]
    \centering
    \subfigure[Adaptive Attack]{
		\includegraphics[width=0.4\linewidth]{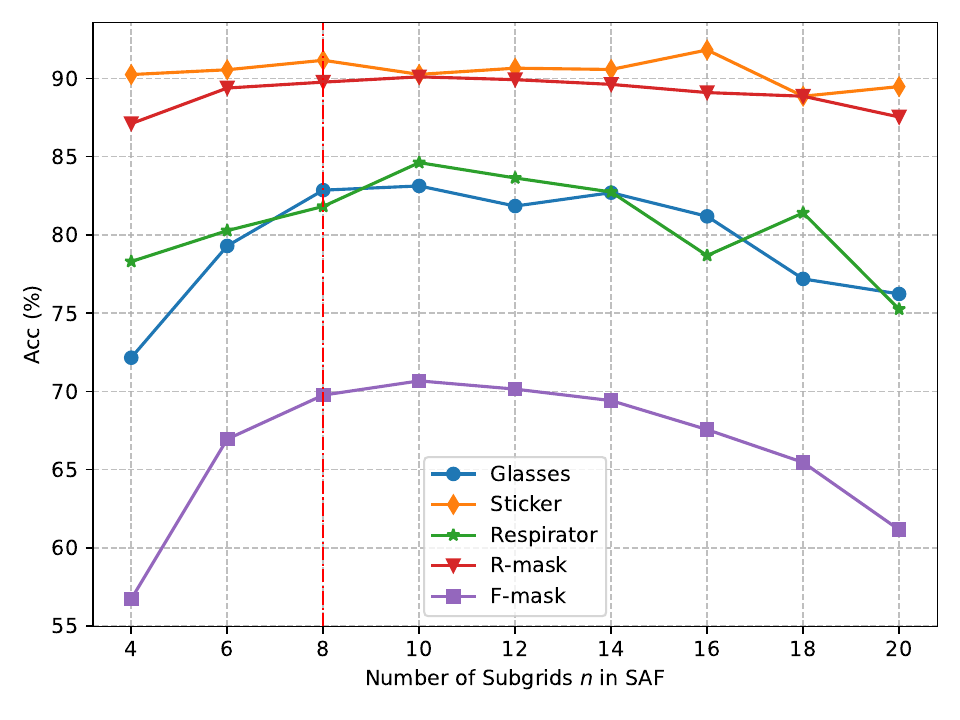}}
    \subfigure[Non-adaptive Attack]{
		\includegraphics[width=0.4\linewidth]{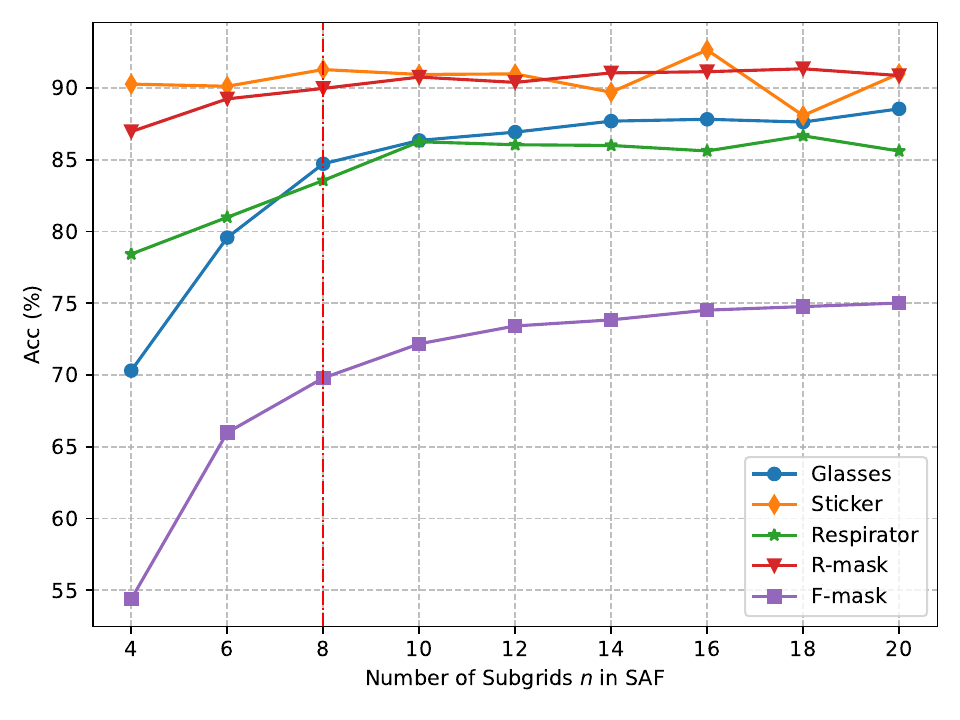}}
    \caption{Number of subgrids ($n$) in SAF. }
    \label{fig:n}
\end{figure*}

To assess the defense method's robustness, we conduct evaluations under adaptive attacks, considering scenarios where the attackers possess comprehensive knowledge about the FR systems due to defense method leakage. In our experiments, we employ the BPDA adaptive attack, which employs a differentiable operation to approximate the non-differentiable component of the original pipeline, enabling adaptive attacks. For ``JPEG" and ``LGS", we assume the output of each defense is approximately equal to the original input. In the case of ``SAC" and ``\textbf{Ours}", we utilize a Sigmoid function to approximate the non-differentiable binarization operation. The results for defense against adaptive attacks are presented in Table~\ref{tab:bpda}. The table clearly indicates that ``JPEG'', ``LGS'', and ``\textbf{Ours($-$)}'' exhibit weak defense against adaptive attacks, while ``\textbf{Ours($+$)}'' demonstrates the highest defense performance across all patch masks.

Figure~\ref{fig:non-a-c} provides a comparative analysis of defense results against non-adaptive and adaptive attacks.  it is clear that the ``DOA'' method, based on adversarial training, and the ``PC'' defense method, which relies on two-stage judgments, do not suffer from gradient obfuscation issues. Consequently, they exhibit consistent defense performance under both adaptive and non-adaptive attacks. ``\textbf{Ours($+$)}'' also maintains similar defense performance in both scenarios, highlighting the robustness of our defense method.

Figure~\ref{fig:A_attack} offers visual results of the defense masks under adaptive attacks. The ``Glasses" example in Figure~\ref{fig:A_attack} illustrates SAC's subpar defense when facing patches with complex shapes, and the ``F-mask" demonstrates SAC's inadequacy when dealing with patches featuring random multiple holes. In contrast, ``\textbf{Ours($+$)}'' effectively defends against both cases, showcasing the superior defense robustness and adaptability of our method.

In Figure~\ref{fig:n}, we present defense results at different numbers of subgrids ($n$). The figure indicates that the defense against adaptive attacks is influenced by the hyperparameter $n$. When $n$ is too small, it tends to over-mask patches, while when $n$ is too large, it under-masks patches. In our experiment, we set $n$ to 8, but practical applications may require adjusting the hyperparameter $n$ to optimize performance.

\section{Discussion}
\label{sec:discussion}
\textbf{Different Patch Shapes.} PC~\cite{xiang2022patchcleanser} effectively defends against small patches but struggles with larger ones. Both PC and SAC~\cite{liu2022segment} face challenges when dealing with intricate patch shapes. To enhance our method’s robustness and adaptability, we propose F-patch, which includes a variety of shapes for training the patch segmenter and the SAF strategy, enabling adaptive matching of patch shapes against patches of varying shapes.

\textbf{Multiple Patch Holes.} While both PC and SAC have mechanisms to address scenarios involving multiple patch holes, practical challenges arise when attempting to determine the exact number of patch holes employed by an attacker. As a result, their defense strategies prove less effective when dealing with situations involving random multiple patch holes. In contrast, our approach excels in providing adaptive defense in such complex cases.

\textbf{Obfuscated Gradients.} Similar to JPEG~\cite{dziugaite2016study} and LGS~\cite{naseer2019local}, our method without SAF suffers from the problem of obfuscating the gradient. To mitigate this problem, we introduce the SAF strategy, which significantly enhances our defense method’s robustness.

\textbf{Limitation and Future Work.} It is important to note that compared to theoretically provable defense methods like PC, the inclusion of a deep neural network structural patch segmenter in our approach makes theoretical proof challenging. This limitation points to an area for potential future work.

\section{Conclusions}
\label{sec:conclusions}
In conclusion, our work addresses the critical issue of defending against adversarial patches in both closed-set and open-set FR systems. By introducing the RADAP method, we have presented a robust and adaptive defense approach capable of countering a wide range of adversarial patches. Our proposed techniques, FCutout and F-patch, utilizing randomly sampled masks in the Fourier space, significantly enhance the occlusion robustness of the FR model while optimizing the performance of the patch segmenter. Additionally, the introduction of the EBCE loss function further refines the patch segmenter's capabilities. To fortify our defense strategy against complete white-box adaptive attacks following the patch segmenter model's leakage, we have introduced the SAF strategy.

The results of our extensive experiments not only validate the effectiveness of RADAP but also showcase its substantial improvements in defense performance against diverse adversarial patches. In fact, RADAP outperforms other state-of-the-art methods and even exhibits a higher clean accuracy compared to the undefended Vanilla model.

Our contributions underscore the significance of considering both closed-set and open-set systems in developing defense strategies for real-world FR applications. As adversarial threats continue to evolve, our work stands as a crucial step forward in enhancing the security and robustness of FR systems against adversarial attacks. We hope that our efforts will inspire further research in this vital domain and foster the development of even more resilient defense mechanisms.

\section*{Acknowledgments}
This work was supported in part by the STI 2030-Major Projects of China under Grant 2021ZD0201300, and by the National Science Foundation of China under Grant 62276127.
 \bibliographystyle{elsarticle-num} 
 \bibliography{cas-refs}





\end{document}